\documentclass[a4paper,fleqn]{cas-dc}
\usepackage[numbers,sort&compress]{natbib}
\usepackage{graphicx}
\usepackage{subcaption}

\begin{document}
\shorttitle{Flow fields prediction based on large language models}

\title [mode = title]{FlowBERT: Prompt-tuned BERT for variable flow field prediction}



\author[1]{Weihao Zou}[type=author,
                        auid=000,
                        orcid=0009-0008-1706-3311]
\author[1]{Weibing Feng}
\cormark[1]
\ead{wbfeng@shu.edu.cn}
\author[1]{Pin Wu}
\cormark[1]
\ead{wupin@shu.edu.cn}
\author[1]{Jiangnan Wu}
\author[1]{Yiguo Yang}

\address[1]{School of Computer Engineering and Science, Shanghai University, Shanghai 200444, China}

\begin{abstract}
  This study proposes a universal flow field prediction framework based on knowledge transfer 
  from large language model (LLM), addressing the high computational costs of traditional 
  computational fluid dynamics (CFD) methods and the limited cross-condition transfer capability 
  of existing deep learning models. The framework innovatively integrates Proper Orthogonal 
  Decomposition (POD) dimensionality reduction with fine-tuning strategies for pretrained LLM, 
  where POD facilitates compressed representation of flow field features while the fine-tuned model 
  learns to encode system dynamics in state space. To enhance the model's adaptability to flow field 
  data, we specifically designed fluid dynamics-oriented text templates that improve predictive 
  performance through enriched contextual semantic information. Experimental results demonstrate 
  that our framework outperforms conventional Transformer models in few-shot learning scenarios while 
  exhibiting exceptional generalization across various inflow conditions and airfoil geometries. 
  Ablation studies reveal the contributions of key components in the FlowBERT architecture. Compared 
  to traditional Navier-Stokes equation solvers requiring hours of computation, our approach reduces
  prediction time to seconds while maintaining over 90\% accuracy. The developed knowledge transfer 
  paradigm establishes a new direction for rapid fluid dynamics prediction, with potential 
  applications extending to aerodynamic optimization, flow control, and other engineering domains.
\end{abstract}










\maketitle

\section{Introduction}
\vspace{1em}

The continuous advancement of computational technology and the refinement of fluid 
mechanics theory have significantly expanded the application scope of Computational 
Fluid Dynamics (CFD). Researchers leverage CFD's precise numerical simulation capabilities 
to investigate intricate details of diverse flow phenomena \cite{liu2021experimental,wang2023tanh,rahimzadeh2023development}. When addressing 
complex fluid mechanics challenges, ensuring simulation accuracy necessitates meticulous 
grid generation, advanced Navier-Stokes equation solving techniques, and precise handling 
of dynamic mesh variations. These computationally intensive tasks inevitably elevate costs, 
thereby prolonging structural design optimization cycles \cite{hou2022novel}. In practical engineering 
applications, increasingly complex computational domains lead to exponential growth in 
grid node counts. Current CFD solvers for airfoil flow field analysis require extensive 
iterative processes, which demand substantial storage resources and prove computationally 
expensive and time-consuming. Reduced-order methods such as Proper Orthogonal Decomposition 
\cite{dowell1997eigenmode,Yang10032025} and Dynamic Mode Decomposition (DMD) \cite{schmid2010dynamic} have markedly reduced the complexity of 
solving such systems while enhancing modeling efficiency.

Numerical simulation serves as a critical predictive tool for unraveling complex physical 
processes, yet its computational resource demands remain prohibitive despite delivering 
precise results. Reduced-Order Model (ROM) address this by simplifying computational 
systems, enabling efficient simulations under limited resources while preserving reliability \cite{schilders2008model}.
ROMs have found widespread application in CFD across environmental science, aerospace, 
and industrial domains \cite{cao2006reduced, xiao2019machine, cstefuanescu2013pod, lieu2006reduced, hesse2014reduced}.

Reduced-order model is a technique that approximates the temporal evolution of physical 
systems by capturing their coherent modes and structures \cite{lucia2004reduced}. Its core typically consists of a 
dimensionality reduction mechanism and a dynamical model in the reduced-order state space.

Research on dimensionality reduction mechanisms has advanced significantly since POD was first successfully applied to ROM construction in 
1996. Numerous POD-based ROM approaches have since been developed, falling into two main categories: 
intrusive and non-intrusive methods. The classical intrusive ROM combines POD with Galerkin projection 
techniques, while non-intrusive variants typically integrate POD with interpolation methods such as 
Kriging, Radial Basis Functions (RBF), and SNORAK - their predictive accuracy being largely determined 
by the choice of interpolation functions and the quality of sample data. Recent advances in machine 
learning and deep learning have introduced new techniques like Dynamic Mode Decomposition (DMD) \cite{takeishi2017learning, li2022linear} 
and autoencoders into ROM development. Notable applications include Eivazi et al.'s \cite{eivazi2020deep, lee2020model} 
autoencoder-based non-intrusive ROM and Kou et al.'s \cite{kou2018reduced} DMD approach for flow field feature 
extraction. However, the complex architectures and numerous parameters of autoencoders often lead to 
training difficulties and overfitting, making traditional methods like POD and DMD preferable in 
certain scenarios. While POD remains a fundamental linear dimensionality reduction technique in fluid 
dynamics through its projection of high-dimensional states onto low-dimensional subspaces, machine 
learning approaches particularly convolutional autoencoders (CAE) have demonstrated remarkable 
progress in efficient nonlinear reduction \cite{lusch2018deep, pan2023neural}, especially for fluid flow dimensionality reduction 
and feature extraction \cite{murata2020nonlinear, maulik2021reduced}. Although POD and DMD maintain their importance in ROM, the integration 
of deep learning has expanded the field's possibilities and driven its advancement. This work focuses 
on exploring fine-tuned large model approaches for dynamical modeling in reduced-order state spaces, 
leading to our selection of POD for the model reduction component.

The investigation of dynamical models in reduced-order state spaces remains equally critical. With the 
advancement of deep learning technologies, neural network-based ROM construction has emerged as a 
prominent research focus. Deep learning excels at uncovering complex data relationships and possesses 
exceptional function approximation capabilities, theoretically enabling bounded functions to be 
approximated with arbitrary accuracy in any space \cite{hornik1989multilayer}. This remarkable property makes neural networks 
particularly suitable for simulating intricate flow phenomena. A growing body of research demonstrates 
successful integration of ROM frameworks with deep learning methodologies, with findings consistently 
indicating that such hybrid approaches significantly enhance ROM's potential and broaden its application 
prospects.

In non-intrusive ROM development, deep learning primarily addresses time-series modeling—traditionally 
reliant on interpolation techniques. Recurrent Neural Networks (RNN), particularly Long 
Short-Term Memory (LSTM) networks \cite{hochreiter1997long}, excel at sequence modeling. Consequently, integrating 
RNNs with POD has emerged as a key ROM development direction. Wang et al. \cite{wang2018model} proposed a 
POD-LSTM Deep Learning ROM that effectively captures fluid flow complexity. Temporal 
Convolutional Networks (TCN) also demonstrate strong sequence modeling capabilities, with 
Wu et al. \cite{wu2020data} developing a streamlined POD-TCN ROM.

Despite RNNs' sequence modeling strengths, their intricate architectures, parallelization 
challenges, gradient vanishing issues, and long-range dependency limitations restrict broader 
application. Similarly, Convolutional Neural Networks (CNN) struggle to capture global flow 
variations due to localized receptive fields. Transformer networks \cite{vaswani2017attention} overcome these 
limitations through self-attention mechanisms, enabling each temporal flow data point to 
dynamically weight information across the entire sequence. Wu et al. \cite{wu2022non} demonstrated that 
POD-Transformer ROMs reduce prediction errors by 35\% and 60\% compared to LSTM and CNN 
counterparts. Ami et al. \cite{hemmasian2023reduced} employed convolutional autoencoders for dimensionality reduction 
coupled with Transformer-based dynamics modeling, maintaining competitive performance even on 
turbulent flow datasets. Nevertheless, predicting flow fields under diverse conditions still 
requires extensive data to train multiple specialized models.

To address the challenges of high computational costs in data acquisition and model training, 
an effective methodology is required for learning from limited training data while achieving 
accurate flow field predictions across diverse conditions. Recent advances in large-scale 
models offer promising solutions. Large Language Models have demonstrated remarkable 
success in computer vision and natural language processing, and this technology is now being 
extended to time-series prediction. LLM exhibit exceptional generalization capabilities, 
efficient data utilization, sophisticated reasoning skills, and multimodal knowledge integration,
making them particularly suitable for predictive applications.
In the field of fluid mechanics, Du et al. \cite{du2024large} proposed an LLM-based intelligent equation 
discovery framework that generates and optimizes governing equations through natural language 
prompts. This approach achieves physically interpretable automatic equation discovery in various 
nonlinear systems, outperforming traditional symbolic mathematics methods. Similarly, 
Pan et al. \cite{pandey2025openfoamgpt} developed the OpenFOAM-GPT agent, which integrates GPT-4 and o1 models to enable 
intelligent OpenFOAM simulation processing. Their system demonstrates outstanding zero-shot 
adaptation capabilities and engineering practicality for complex CFD tasks.

Unlike traditional prediction methods, LLM eliminate the need for comprehensive retraining for 
each new task. They can achieve accurate predictions with minimal data and simplified optimization 
processes, suggesting significant potential for applying LLM to time-series data analysis. 
This has led to the emergence of various large-scale models such as BERT \cite{devlin2019bert} and GPT-2 \cite{radford2019language}. 
When Jin et al. \cite{jin2023time} and Zhou et al. \cite{zhou2023one} fine-tuned pre-trained large models for time-series 
tasks, they demonstrated exceptional performance, confirming the applicability of large models 
to temporal problems and laying the foundation for their use in time-dependent flow field prediction.
From a large model perspective, BERT represents a landmark pre-trained language model that has 
demonstrated outstanding performance in large-scale Natural Language Processing (NLP) tasks. 
Employing a bidirectional Transformer architecture, BERT processes each word while simultaneously 
considering both left and right contextual information. This bidirectional contextual understanding 
significantly enhances the model's language comprehension capabilities, which similarly provides a 
foundation for understanding temporal evolution in flow fields.
Consequently, we selected BERT-based fine-tuning as our dynamical modeling approach in reduced-order 
state space. This method implicitly incorporates expert knowledge and physical principles into the 
reduced-order model through knowledge distillation and few-shot learning strategies, enabling the model 
to maintain strong performance even with limited training data.

Traditional reduced-order models have demonstrated excellent performance in fitting single flow 
field datasets to meet required specifications. However, a significant limitation persists: different flow 
conditions typically necessitate training separate models to achieve comparable accuracy, substantially 
increasing data requirements. To address these challenges, this paper proposes a novel ROM framework that 
integrates POD with pre-trained large language models.
Our methodology employs CFD-generated flow field snapshots to construct the dataset, where POD extracts 
basis functions to generate optimal solution representations, while fine-tuned pre-trained large models 
perform time-series prediction. 

We developed a specialized text templating system coupled with temporal 
text embedding technology, which implicitly incorporates textual information into flow field data through 
embedding techniques. The processed text templates are then aligned with these vectors using language models 
to facilitate large model fine-tuning.
Benchmark tests on cylinder wake flow demonstrate that the model maintains over 90\% prediction accuracy even 
with 80\% fewer training samples. Ablation studies reveal a consistent inverse relationship between text-template 
compatibility and prediction error - higher template-to-flowfield adaptation yields progressively lower errors. 
Cross-scenario validation shows the model achieves approximately 95\% prediction accuracy on airfoil datasets, 
confirming its superior transfer capability.Key contributions include:

\begin{enumerate}
  \item[1)] A FlowBERT framework integrating pre-trained language model fine-tuning and reduced-order 
  modeling for robust aerodynamic flow field prediction.

  \item[2)] Flow-adapted textual prompting templates enhancing few-shot prediction while reducing data 
  acquisition costs.

  \item[3)] A transfer learning framework generalizable across diverse flow conditions, significantly 
  lowering modeling expenses.
\end{enumerate}

The paper is structured as follows: Section II details methodological foundations. Section III 
presents the framework architecture. Section IV discusses experimental design and analysis. 
Conclusions are summarized in Section V.

\section{Methodology}
\vspace{1em}

This section introduces several methods and concepts involved in the model: the dimensionality reduction 
method of intrinsic orthogonal decomposition, data normalization methods, the multi-head cross-attention 
mechanism, and the structure of pre-trained large models.

\subsection{Proper orthogonal decomposition}
\vspace{1em}

To obtain a low-dimensional representation of the flow field, one of the most commonly used methods is 
POD. Our ROM employs POD for the dimensionality reduction and reconstruction of the flow field. 
The goal of POD is to determine a sequence of basis functions within the continuous flow field 
snapshot space that can almost entirely capture the information of the flow field snapshots, allowing 
each snapshot to be approximated using this set of basis functions. By extracting this information, 
a low-order representation of the flow field can be achieved.aaa

The calculation process of POD is as follows: first, we consider a parameterized model:
\begin{equation}
  E(\mu)\frac{d\mathbf{u}}{dt} + A(\mu)\mathbf{u} = f(\mu,\mathbf{u}),
\end{equation}
where $\mathbf{u}\in \mathbb{R}^{n}$ , \( A(\mu) \in \mathbb{R}^{n \times n} \), \( \mathbf{E}(\mu) \in \mathbb{R}^{m \times m} \), and \( f(\mu, \mathbf{u}) \in \mathbb{R}^{n} \). We first consider a model where the right-hand side is linear and parameter-independent:
\begin{equation}
  \mathbf{E}(\mu)\frac{d\mathbf{u}}{dt} + A(\mu)\mathbf{u} = b.
\end{equation}

We take numerical method solutions at $n_{s}$ time points, arranging each time point into a column to form a matrix, as follows:
\begin{equation}
  \mathbf{U}\in R^{n\times n_s},\mathbf{U}=[u_1,u_2,\ldots,u_{n_s}],
\end{equation}
perform singular value decomposition (SVD) on the matrix U as follows:
\begin{equation}
  \mathbf{U}=\mathbf{V}\Sigma \mathbf{W}^{\mathrm{T}},
\end{equation}
by retaining the first $r$ columns of the left singular matrix $\mathbf{V}$, we construct 
the reduced basis matrix $\mathbf{V_{r}}=[v_{1},\ldots,v_{r}]$, which forms the optimal POD basis. 
This basis satisfies the minimum projection error criterion, mathematically 
expressed as:
\begin{equation}
  \begin{aligned}
    \min\limits_{V\in \mathbb{R}^{n\times r}} & \| \mathbf{U} - \mathbf{V}\mathbf{V}^{\mathrm{T}} \mathbf{U} \|_{F}^{2} \\
    = & \min\limits_{V\in \mathbb{R}^{n\times r}} \sum\limits_{i=1}^{n_{s}} \| u_{i} - \mathbf{V}\mathbf{V}^{\mathrm{T}} u_{i} \|_{2}^{2} = \sum\limits_{i=r+1}^{n_{s}} \sigma_{i}^{2},
  \end{aligned}
\end{equation}
we let $\mathbf{u_{r}}=\mathbf{V_{r}}^{T}\mathbf{u}$ and substitute this into the equation. By left-multiplying the final expression by $V_{r}^{T}$, we can obtain a low-order equation expressed as follows:
\begin{equation}
  \mathbf{E_r}\frac{d\mathbf{u_r}}{\mathrm{d}t}+\mathbf{A_r}\mathbf{u_r}=\mathbf{b_r},
\end{equation}
where $\mathbf{E_{r}}=\mathbf{V_{r}}^{T}\mathbf{E}\mathbf{V_{r}}$, $\mathbf{A_{r}}=\mathbf{V_{r}}^{t}\mathbf{A}\mathbf{V_{r}}$, $\mathbf{b_{r}}=\mathbf{V_{r}}^{T}\mathbf{b}$. Thus, the original n-th order model is reduced to an r-th order model, which significantly reduces the computational workload and enhances computational efficiency in subsequent reasoning processes.

\subsection{Pre-trained large models}
\vspace{1em}

\begin{figure}
	\centering
	\includegraphics[width=0.8\columnwidth]{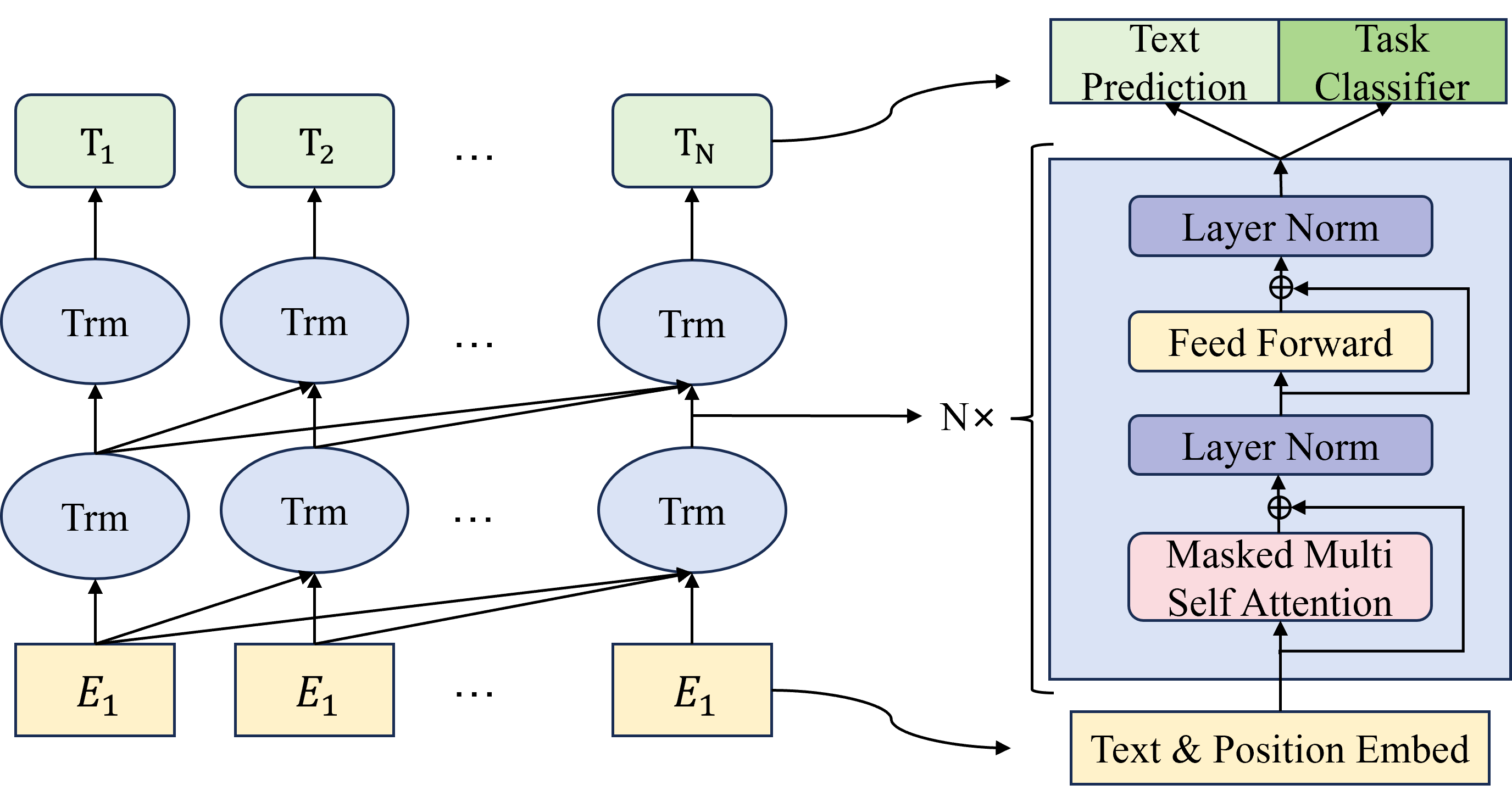}
	\caption{Pre-trained large model.}
	\label{FIG.1}
\end{figure}

With the continuous advancement of deep learning technology, pre-trained large models have achieved 
significant results in fields such as natural language processing and computer vision. By being pre-trained 
on massive datasets, these models can capture rich semantic information, providing powerful feature 
representations for downstream tasks.

During the pre-training phase, data undergoes preprocessing, including tokenization and encoding. 
The preprocessed data serves as input to the model. Pre-trained large models typically utilize deep 
neural network architectures, such as Transformers. The Transformer model features a self-attention 
mechanism that captures long-range dependencies in sequential data. 
Within each Transformer structure, the encoder maps a set of inputs $x=\begin{bmatrix}x_1,...,x_n\end{bmatrix}$ 
to a set of output sequences $z=\begin{bmatrix}z_1,...,z_n\end{bmatrix}$. For a given z,the decoder 
iteratively outputs values $y=\begin{bmatrix}y_1,...,y_n\end{bmatrix}$.
Fig. {\ref{FIG.1}} illustrates the structure of the pre-trained large model. Pre-training occurs through 
two main tasks: (1) Masked Language Model (MLM): randomly masking certain words in the input sequence, 
prompting the model to predict these masked words; (2) Next Sentence Prediction (NSP): given two sentences, 
the model predicts whether the second sentence follows the first. 
Through these pre-training tasks, the model learns rich semantic information.

After completing pre-training, pre-trained large models can be applied to various downstream tasks, 
such as text classification, sentiment analysis, and named entity recognition. For specific downstream 
tasks, some parameters of the pre-trained model are fine-tuned to adapt the model to the new task. 
Additionally, the pre-trained model can serve as a feature extractor, extracting feature representations 
from input data, which can then be fed into other models for training.

\subsection{Multi-head Cross-attention Mechanism}
\vspace{1em}

\begin{figure}
	\centering
	\includegraphics[width=0.6\columnwidth]{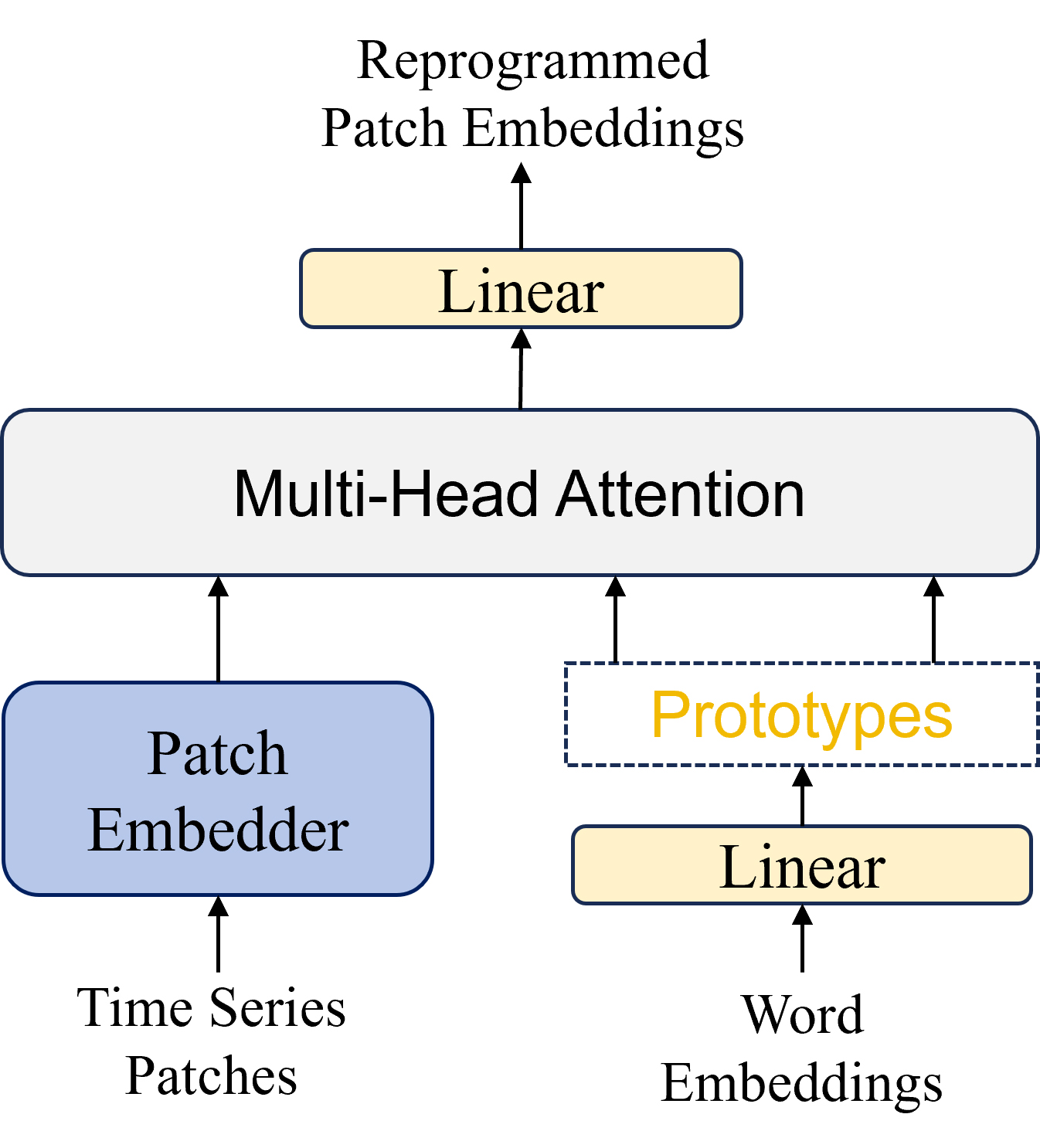}
	\caption{Multi-head cross-attention mechanism.}
	\label{FIG.2}
\end{figure}

The multi-head attention mechanism is a key component for processing sequential data, allowing the model 
to focus on important parts of the input by assigning different attention weights. In Transformer-based 
models, the attention mechanism is divided into multiple heads, each with its own parameters, capturing 
information from different subsets. The attention mechanism computes the output based on Query, Key, and  
Value, with the weights determined by the interaction between the Query and the Key, which are then used 
to weight the Value to produce the output.In Transformers, the initialization of parameter matrices $W_Q$, 
$W_K$ and $W_V$ assists in the self-attention computation by integrating with the input matrix $\text{Y}$ 
and incorporating positional information.We use multi-head cross-attention to recompile the temporal flow 
field data into text vectors, allowing the flow field data to possess some textual information, thus better 
adapting to the input of pre-trained large models.For a given flow field feature block $x_i\in\mathbb{R}^{D\times d_m}$, 
we first compute the cross-attention with the text prototype $x_{emb}\in\mathbb{R}^{V^{\prime}\times d_{m}}$ to 
establish a relationship with the text prototype. The formula for cross-attention computation is as follows:
\begin{equation}
  \begin{gathered}
    \mathrm{Attention}(\mathbf{Q}, \mathbf{K}, \mathbf{V}) = \mathrm{softmax}\left(\frac{\mathbf{Q}\mathbf{K}^{\top}}{\sqrt{\mathbf{d_{k}}}}\right)\mathbf{V}, \\
    \mathrm{Cattention}(\mathbf{x_{i}}, \mathbf{x_{emb}}) = \mathrm{Attention}(\mathbf{Q_{x_i}}, \mathbf{K_{x_{\mathrm{emb}}}}, \mathbf{V_{x_{\mathrm{emb}}}}), 
  \end{gathered}
\end{equation}
here ${\mathbf{Q},\mathbf{K},\mathbf{V}}\in\mathbb{R}^{D\times d_{m}}$ represent the query, key, and value matrices, 
respectively, D denotes the number of patches, and $d_{m}$ indicates the dimensionality of the embeddings. 
The cross-attention scores for individual patch blocks are computed using the CrossAttention function.
the higher the correlation between the data block and the text prototype, the greater the corresponding 
attention weight. Multi-head cross-attention is an extension of cross-attention, as shown in Fig. {\ref{FIG.2}}
Its computation process is consistent with that of cross-attention. The final output is achieved by merging 
multiple independent cross-attention outputs, calculated as follows:
\begin{equation}
  \mathbf{O}=\mathrm{Linear}(Concat(\mathrm{Cattention_{1}},...,\mathrm{Cattention_{h}})),
\end{equation}
where the multi-head attention output $\mathbf{O}\in\mathbb{R}^{D\times d_{m}}$ has the 
same dimensions as $\mathrm{Attention_{1}}$ and $\mathrm{Attention_{h}}$. The term Concat denotes the 
concatenation of the cross-attention matrices, while Linear refers to the feedforward neural network in 
the equation.

\section{Model architectures}
\vspace{1em}

\begin{figure*}
	\centering
	\includegraphics[width=0.85\textwidth]{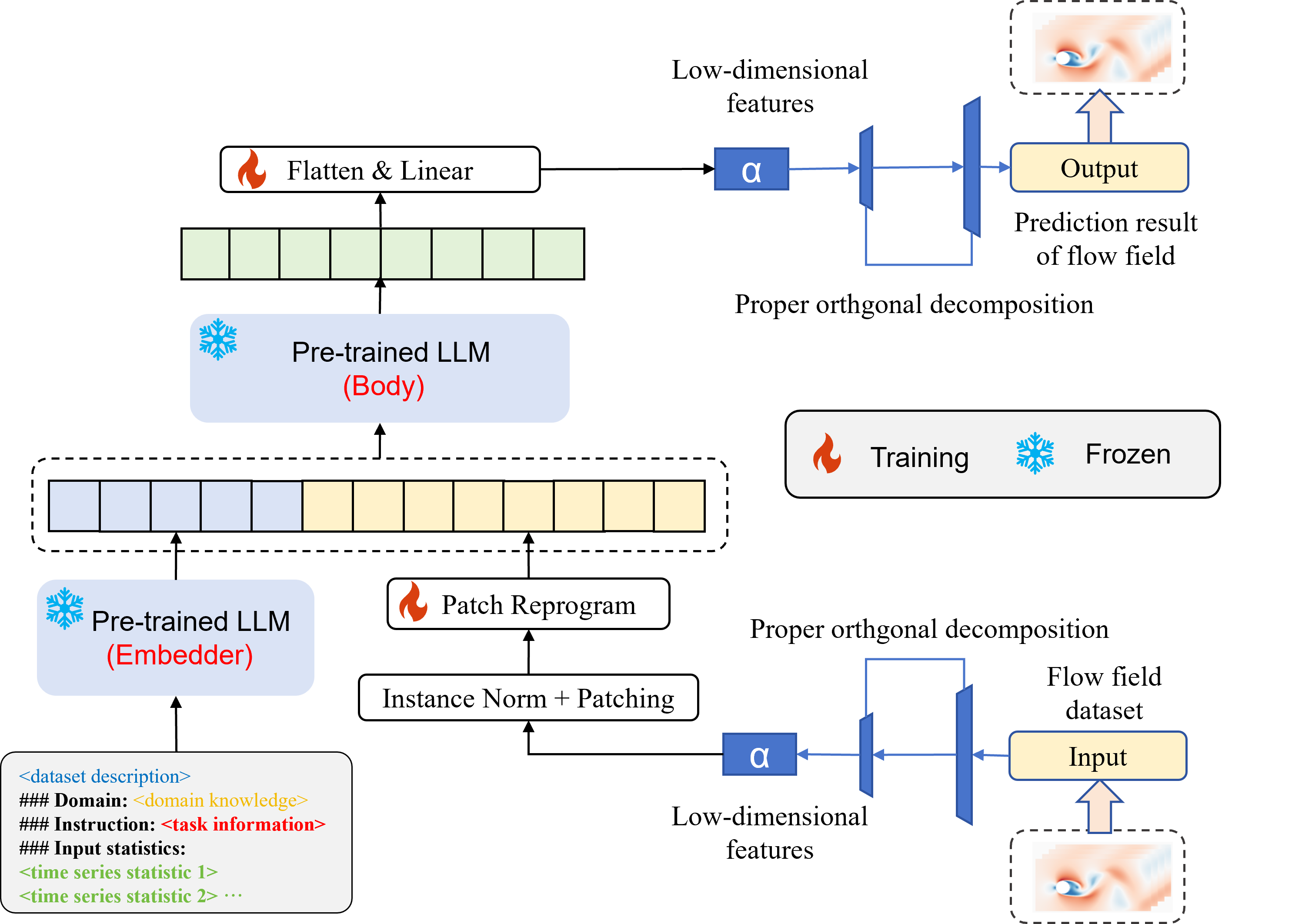}
	\caption{Overall model architecture.}
	\label{FIG.3}
\end{figure*}

The challenge of constructing a flow field prediction model for multiple scenarios lies in the fact that 
previous models excelled only on single, large datasets of flow field data. When transferred to different 
flow field datasets, they struggled to adapt, necessitating a complete retraining of the model, which is 
often unacceptable. To address this issue, we developed the FlowBERT model, which incorporates pre-trained 
large models, allowing the model to achieve excellent performance with only a small amount of training data 
and demonstrating outstanding performance in transfer learning. 
Our model first uses POD to reduce the dimensionality of the flow field data, obtaining low-dimensional 
flow field data. Then, it employs the large model to predict the low-dimensional flow field data, and 
finally retrieves the high-dimensional flow field data through POD.

The overall architecture of our model is shown in Fig. {\ref{FIG.3}}. The model primarily consists of 
four components: the flow field dimensionality reduction method (POD), temporal text embeddings, text 
prompt templates, and the pre-trained large model. The details of each module will be elaborated upon below.

\vspace{1em}

\noindent\textbf{Data dimensionality reduction: }When inputting flow field data, we first apply POD to reduce the dimensionality of 
the flow field data, obtaining low-dimensional representations. This data is then input into the model for 
temporal inference, significantly reducing the computational load and enhancing computational efficiency. 
After the model outputs the results, we use POD to recover the data, resulting in high-dimensional flow 
field data.

\vspace{1em}

\noindent\textbf{Input embedding: }We apply invertible instance normalization to each input 
channel $\mathrm{x}^{(i)}$ to ensure that the data has a mean of zero and a standard deviation of one, 
mitigating distribution shifts in the time series.Subsequently, we slice $\mathrm{x}^{(i)}$ into multiple 
segments of length L, which may overlap or be completely non-overlapping. 
The total number of these input segments is calculated as $D=\lfloor\frac{(\mathrm{T-L})}{S}\rfloor+1$, 
where S is the sliding distance between segments. This process serves a dual purpose: first, to better 
preserve local contextual information by aggregating local details within each segment; and second, to 
construct a concise sequence of input tokens to reduce computational resource consumption.
These segments $\mathbf{X}_D^{(i)}\in\mathbb{R}^{D \times L}$ undergo embedding transformation through 
a linear layer, generating ${\widehat{\mathbf{X}}}_{D}^{(i)}\in\mathbb{R}^{D\times d_{m}}$, where $d_{m}$ 
denotes the dimensionality of the embedding space. 

\vspace{1em}

\begin{figure}
	\centering
	\includegraphics[width=.9\columnwidth]{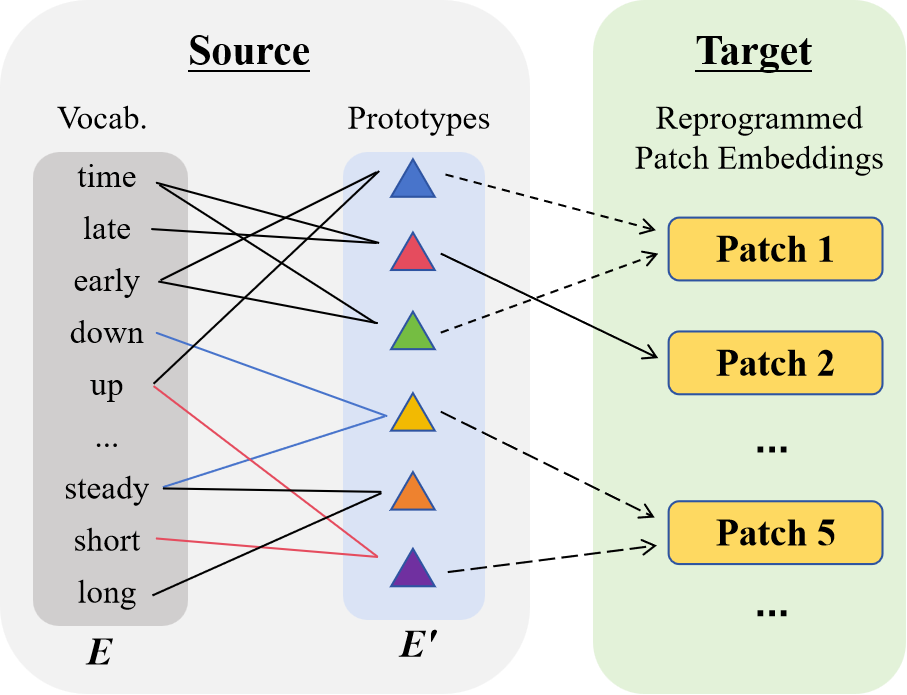}
	\caption{Patch reprogramming.}
	\label{FIG.4}
\end{figure}
\noindent\textbf{Patch reprogramming: }Here, we re-encode the patch embeddings to fit the space of the 
original data representation, aiming to harmonize the flow field time series with natural language patterns, thereby enhancing the backbone network's understanding and reasoning capabilities regarding the time series.
We propose using pre-trained word embeddings $\mathbf{E}\in\mathbb{R}^{V\times d_m}$ to re-encode 
$\mathbf{X}_P^{(i)}$ , where ${V}$ represents the size of the vocabulary. However, due to the lack of 
prior knowledge about which source symbols are directly relevant, directly applying $\mathbf{E}$ could 
lead to a large and potentially complex re-encoding space.  
To address this issue, we adopt a strategy of retaining only a smaller set of text prototypes 
$\mathbf{E}'\in\mathbb{R}^{V^{\prime}\times d_{m}}$ through linear probing, where $V^{\prime}\ll V$. 
Fig. {\ref{FIG.4}} illustrates an example of this concept. The text prototype learning method can 
associate linguistic cues, such as "short up" (red line) and "steady down" (blue line), and integrate 
them to express local patch information (for instance, using "short up followed by steady down" to 
describe patch 5), while avoiding the limitations of pre-trained language models \cite{jin2023time}.
This method is both effective and flexible in selecting relevant source information. To achieve this, 
we apply the multi-head cross-attention mechanism mentioned in the previous section.

\vspace{1em}

\noindent\textbf{Text prompt templates: }Prompts serve as a direct and effective method for activating 
task-specific functionalities of LLM \cite{nie2022time}. However, directly 
converting time series data into natural language is a highly challenging task, which not only hinders 
the creation of automated dataset tracking but also limits the effective use of real-time prompts without 
compromising model performance \cite{cui2024survey}.
Recent studies have shown that other data modalities, such as images, can be seamlessly integrated as 
prompt prefixes, thereby facilitating the reasoning process based on these inputs \cite{xue2022prompt}. 
Inspired by these studies and to ensure our approach can be directly applied to time series in the real 
world, we propose the concept of using prompts as prefixes to enrich the contextual information of the 
input and guide the conversion of time series patches. We found that this method significantly enhances 
the adaptability of LLM in downstream tasks.

\begin{figure}
	\centering
	\includegraphics[width=\columnwidth]{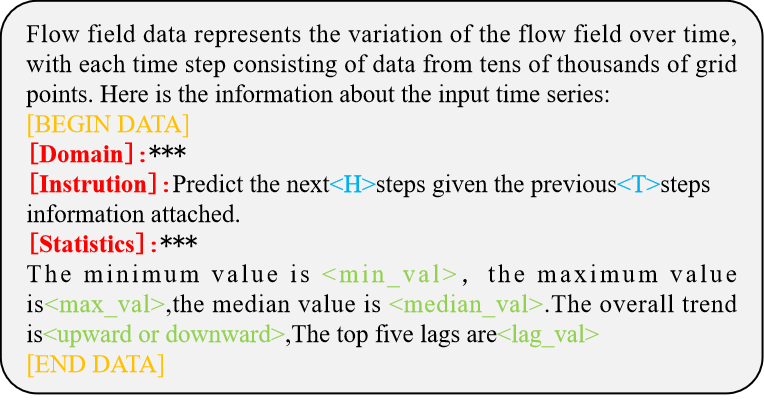}
	\caption{Text prompt templates.}
	\label{FIG.5}
\end{figure}

In the implementation process, we identified three key elements for constructing effective prompts: 
dataset context, task guidance, and input statistics. Fig. {\ref{FIG.5}} presents a template, 
Specific examples will be provided in Section \ref{sec:4}. 
The dataset context utilizes foundational background information about the input time 
series to guide the LLM, which often exhibits unique characteristics across different domains. 
Task guidance serves as an important direction for the LLM in performing patch embedding transformations 
for specific tasks. Additionally, we enhance the input time series by incorporating key statistical 
information to support pattern recognition and reasoning processes.

\vspace{1em}

\noindent\textbf{Output projection: }When processing with a frozen LLM, we remove 
the prefix portion and extract the output representations. Subsequently, we flatten these output 
representations and project them into a low-dimensional space through linear transformations, thereby 
obtaining the final flow field information.

\vspace{1em}

In our model structure, only the parameters for lightweight input transformations and output projections 
are updated, while the backbone language model remains frozen. Compared to other multimodal language models 
which typically fine-tune using paired cross-modal data \cite{tsimpoukelli2021multimodal}, the FlowBERT is 
optimized directly and can be utilized with only a small portion of time series data and a few training 
epochs. This approach maintains high efficiency and imposes fewer resource constraints compared to building 
a domain-specific large model from scratch or fine-tuning an existing one.

\section{Results and discussion}\label{sec:4}
\vspace{1em}

In this section, we evaluate the performance of the proposed ROM through three 
different experiments. First, we examine the model's prediction accuracy in scenarios with limited data 
using the classic two-dimensional flow around a cylinder problem. Second, we conduct transfer learning 
experiments using flow field data from the same type of airfoil to verify the model's transfer learning 
capabilities. The third experiment employs flow field data from three different types of airfoils to explore the model's 
potential and practical effectiveness through transfer learning. All data in these experiments were 
obtained from high-fidelity numerical simulations using structured or unstructured meshes to ensure the 
accuracy of the solutions. 

In the two-dimensional flow around a cylinder case, we collected precise snapshot data over 2000 time 
steps, selecting the first 200 time steps as the training and testing sets for the model. For experiments 
two and three, due to the need for a large amount of high-fidelity data and the high time cost of obtaining 
it, we only collected snapshot data of 110 time steps for each airfoil flow field (from the initial state 
to the converged state). This significantly reduced the time cost of sample acquisition.

We used the Scikit-learn library for singular value decomposition and the PyTorch library to construct the 
neural networks. In all experiments, we employed the root mean square error (RMSE) as the criterion for 
evaluating the accuracy of the ROM. The formula for calculating RMSE is as follows:
\begin{equation}
  \text{RMSE}=\sqrt{\frac{\sum_{i=1}^n\left(X_i^{(t)}-X_{rom,i}^{(t)}\right)^2}{n}},
\end{equation}
where ${n}$ represents the number of nodes in the computational domain, $X_{rom,i}^{(t)}$ is the ROM 
prediction at time step ${i}$, and $X_{i}^{(t)}$ is the corresponding numerical simulation result.

In this work, all experiments were conducted using identical computational hardware: 
an Intel Xeon E5-2678 v3 processor and an NVIDIA GeForce RTX 3090 GPU. The software 
environment included Python 3.9.7, PyTorch 1.11.0, and CUDA Toolkit 11.3.1. For 
model implementation, we employed the pretrained BERT model with the following 
hyperparameters: 20 training epochs, batch size of 12, Gaussian Error Linear Unit 
(GELU) activation functions, mean squared error (MSE) loss function, and the Adam 
optimizer. To enhance training efficiency and reduce resource consumption, we 
integrated DeepSpeed for distributed training optimization.

\subsection{Flow Around a Cylinder Case}
\vspace{1em}


In this subsection, we use the classic two-dimensional flow around a cylinder problem as an example to 
construct the ROM for the velocity vector and validate the performance of the proposed  ROM. 
The computational domain is set as a rectangular area measuring 0.2 meters in length and 0.1 meters 
in width. In this domain, fluid flows in from the left inlet and exits from the right outlet, with a cylinder of 
radius 0.01 meters placed in the middle. We used computational fluid dynamics 
(CFD) numerical simulation software to obtain high-fidelity solutions for the flow field, generating 
high-precision snapshot data over 2000 time steps. To construct the model, 
we selected 10\% of this data, or 200 time steps, as the training and testing sets (with time steps 1 to 
150 used for the training set and time steps 151 to 200 for the testing set).

In the experiment, the fluid velocity is set to v=1 m/s, with a Reynolds number Re = 1000, satisfying 
the condition y += 1. The computational domain consists of 73,851 grid points. Additionally, the time 
window is set to 30 time units. This setup allows us to evaluate the effectiveness and accuracy of the 
ROM in handling flow problems with well-defined physical characteristics.

\begin{figure}
	\centering
	\includegraphics[width=1.1\columnwidth]{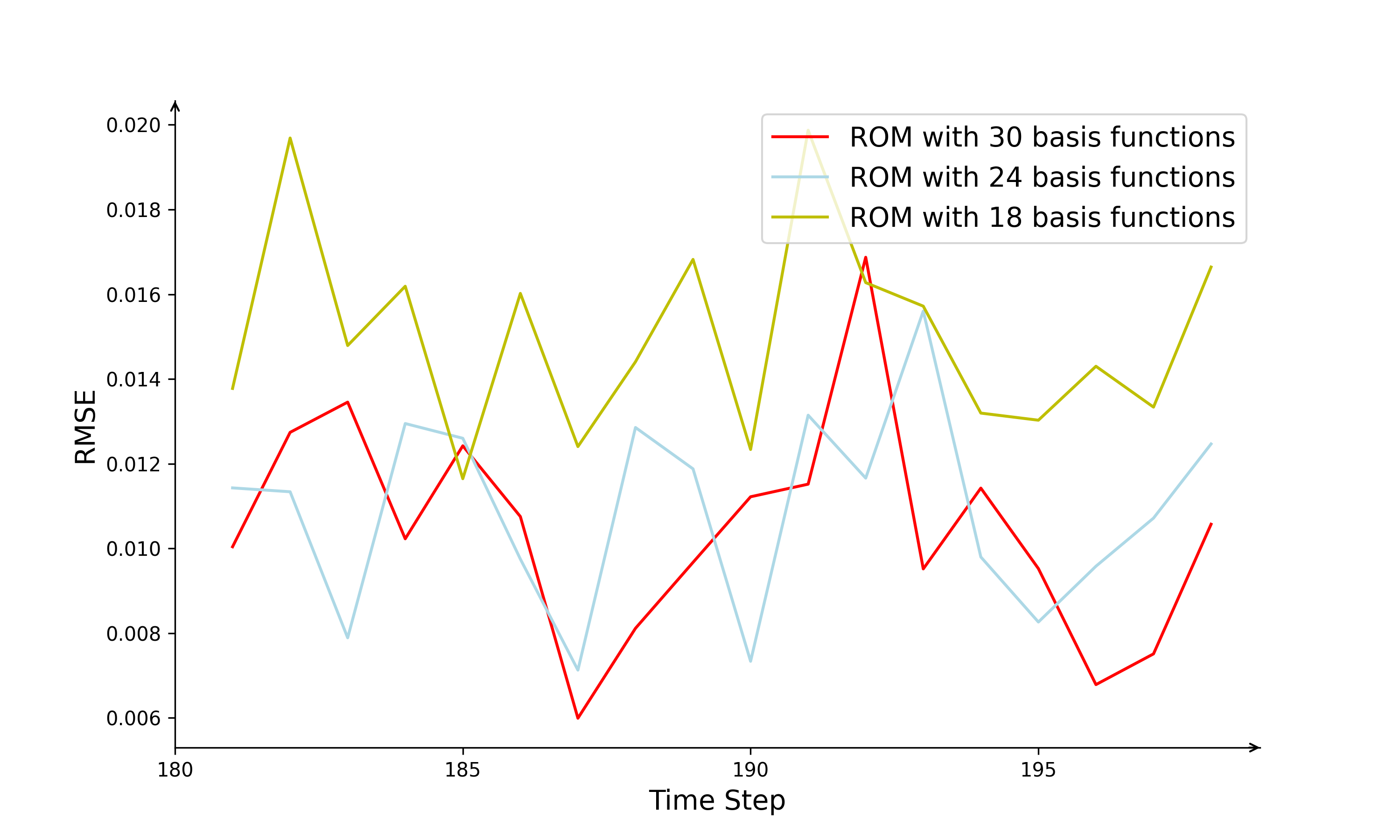}
	\caption{Rmse errors on the validation set with different basis functions.}
	\label{FIG.6}
\end{figure}

To investigate the impact of different numbers of basis functions on model accuracy, we selected three 
different quantities of basis functions (18, 24, and 30) for the experiments. We tested each quantity of 
basis functions and presented the root mean square error (RMSE) curves on the validation set in 
Fig. {\ref{FIG.6}}. The experimental results indicate that as the number of basis functions increases, 
the performance of the ROM improves. This is because more basis functions can capture more information 
from the flow field, thereby reducing the reconstruction error. However, as the number of basis functions 
continues to increase, the rate of error reduction gradually slows down, suggesting that merely increasing 
the number of basis functions does not significantly enhance model performance 
and also leads to increased computational costs. Furthermore, when the number of basis functions is 
excessive, subsequent processing of the low-dimensional flow field data becomes more complex. Therefore, 
in the experiments of this subsection, we consistently used 30 basis functions to balance model 
performance and computational efficiency.

\begin{figure}
	\centering
	\includegraphics[width=\columnwidth]{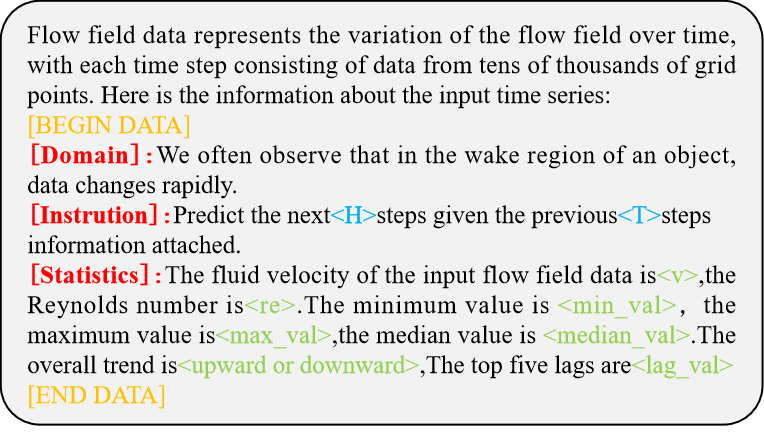}
	\caption{Cylinder flow test prompt templates.}
	\label{FIG.7}
\end{figure}

\begin{figure*}
	\centering
	\includegraphics[width=0.8\textwidth]{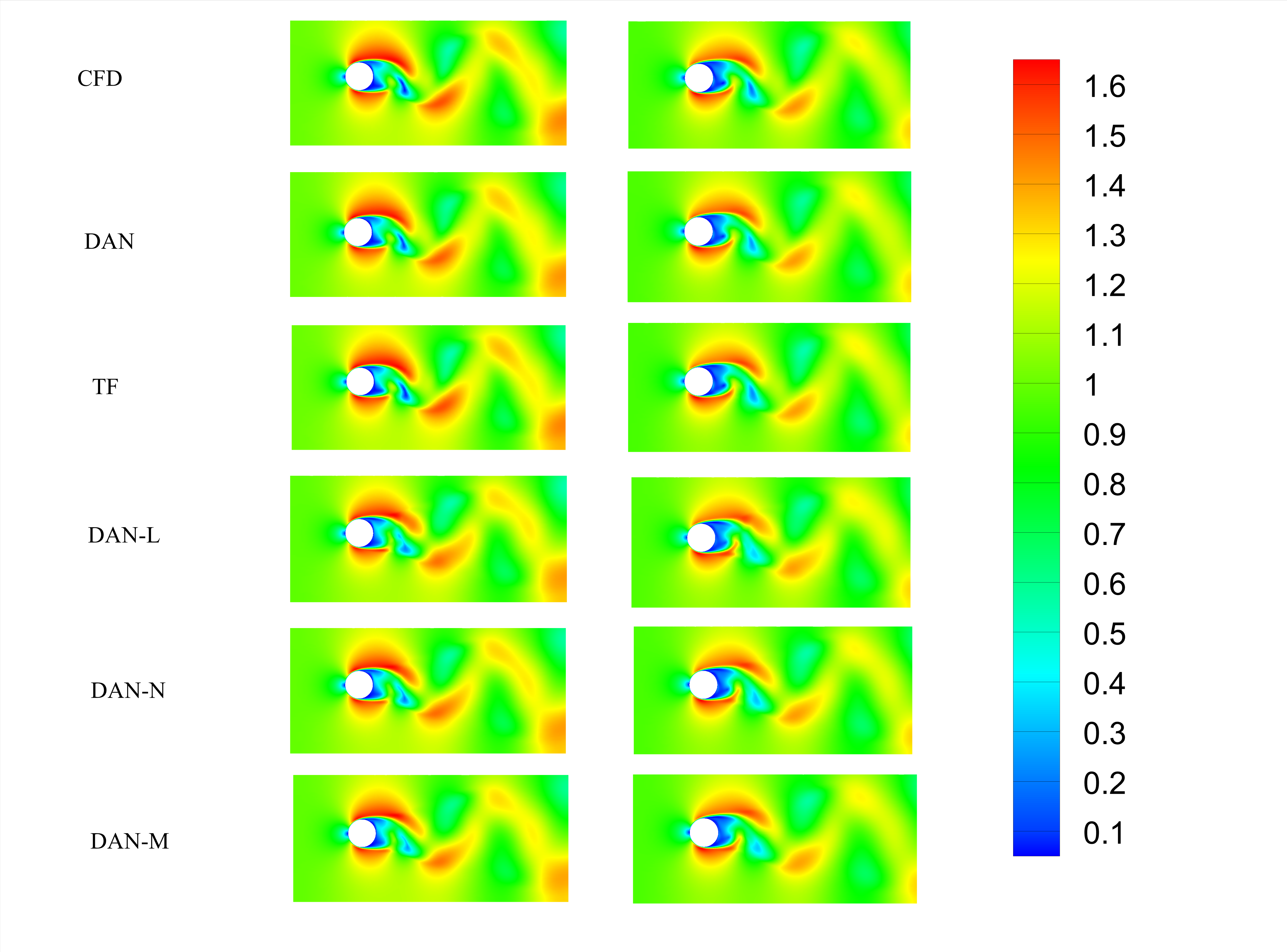}
	\caption{The predictions of the velocity fields for different models at the 180th step (left) and the 190th 
  step (right) (DAN represents our proposed FlowBERT, DAN-L represents FlowBERT with linear layers replacing 
  text embeddings, DAN-N represents FlowBERT without text prompt templates, DAN-M represents FlowBERT with 
  meaningless templates added, and TF represents POD-Transformer).}
	\label{FIG.8}
\end{figure*}

\begin{figure*}
	\centering
	\includegraphics[width=0.8\textwidth]{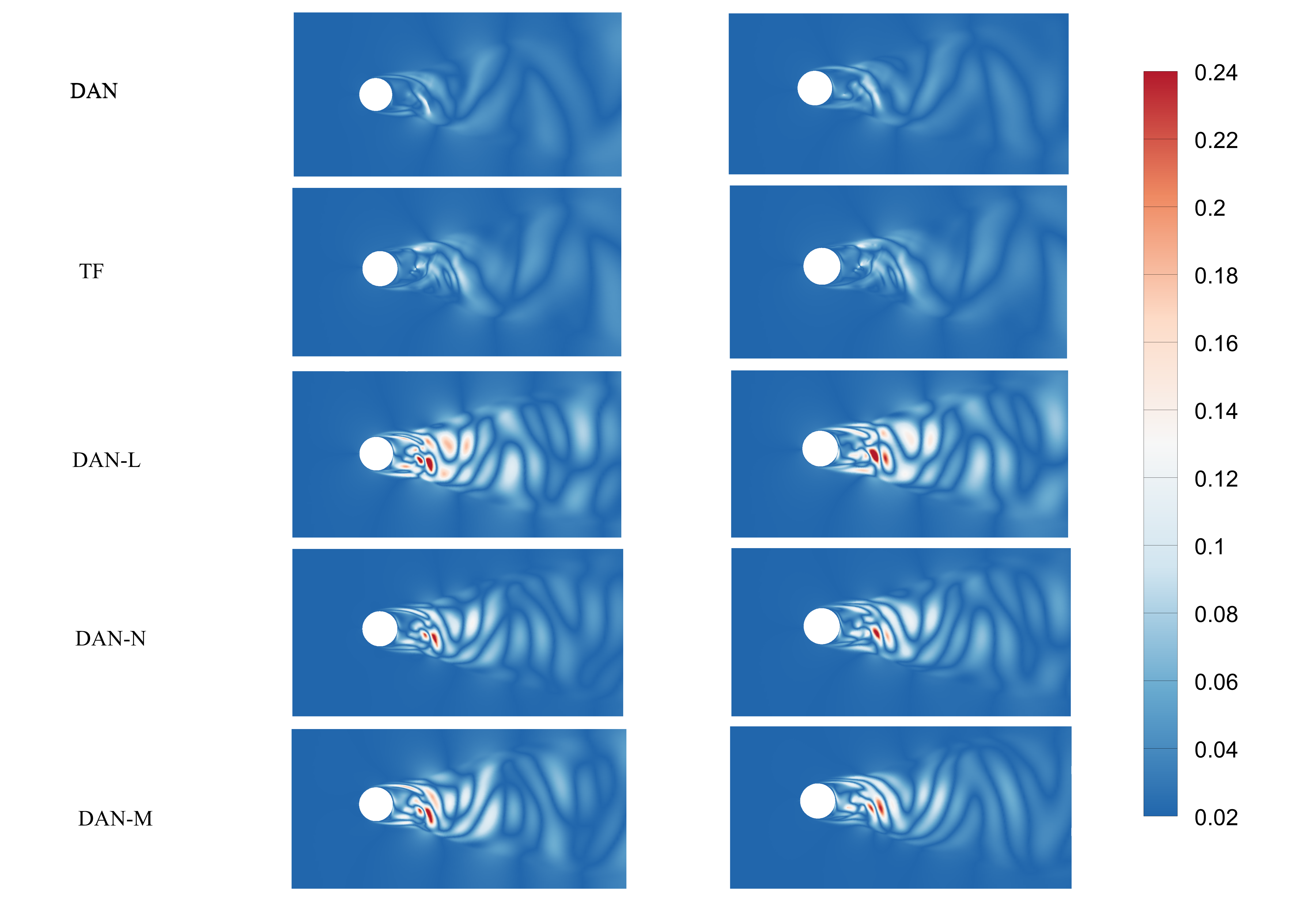}
	\caption{The absolute errors of the velocity fields for different models at the 180th step (left) and 
  the 190th step (right) (DAN represents our proposed FlowBERT, DAN-L represents FlowBERT with linear layers 
  replacing text embeddings, DAN-N represents FlowBERT without text prompt templates, DAN-M represents 
  FlowBERT with meaningless templates added, and TF represents POD-Transformer).}
	\label{FIG.9}
\end{figure*}

In this study, we propose a method utilizing text templates to guide large models for 
flow field prediction, with the text template design for this experiment illustrated in 
Fig. {\ref{FIG.7}}. To validate the effectiveness of this approach, we designed and conducted ablation 
experiments comparing five variants: the complete ROM incorporating text templates, the ROM 
without text templates, the ROM with nonsensical text templates, the ROM replacing text 
embeddings with linear layers, and the Transformer-based ROM under identical conditions. 
As shown in Fig. {\ref{FIG.8}}, the predictions of these five models at the 180th and 190th timesteps 
were compared against CFD simulation results.

The experimental results demonstrate that our complete ROM model achieves higher precision in 
predicting the evolution of flow field details, particularly in capturing near-wall features and 
regions of high nonlinear intensity within the wake. While the template-free ROM, nonsensical-template 
ROM, and linear-layer ROM exhibit acceptable performance in near-wall regions, they display 
imperfections in high nonlinear intensity zones of the wake, though remaining broadly viable. 
This indicates that pretrained large models inherently possess prior knowledge of flow fields, 
enabling effective resolution of nonlinear relationships within the data. Comparative analysis 
reveals that employing meaningful text templates and text embedding processing significantly 
enhances training efficiency. On the other hand, the Transformer-based ROM, leveraging its robust 
attention mechanisms, also delivers strong performance under small sample size conditions.

To better visualize flow field prediction errors, we generated error distribution maps for four 
distinct operating conditions. As shown in Fig. \ref{FIG.9}, the absolute velocity error contours 
at the 180th and 190th timesteps reveal the following findings through error visualization analysis: 
The errors in our proposed Reduced Order Model (ROM) are primarily concentrated in wake regions, 
with the maximum absolute error averaging approximately 0.1, outperforming the Transformer-based ROM. 
In comparison, the template-free ROM exhibits errors about 2.5 times greater than our full ROM, the 
invalid-template ROM shows approximately 3 times higher errors, while the linear-layer ROM demonstrates 
up to 3.5 times larger errors. These results conclusively validate the superior performance of our 
ROM under limited data conditions. The flow field text templates and text embedding modules play 
pivotal roles in error reduction, with template validity proving to be a critical determinant of 
prediction accuracy. Notably, employing invalid templates yields counterproductive effects.

\begin{table}[ht]
  \centering
  \caption{Computational times for numerical simulations and ROM training and extrapolation}
  \begin{tabular}{@{}p{0.35\textwidth} p{0.1\textwidth}@{}}
      \toprule
      Method & Time \\ \midrule
      Numerical Simulation & 8280 s \\ 
      ROM Training Time & 900 s \\ 
      Transformer Training Time & 1200 s \\ 
      \bottomrule
  \end{tabular}
  \label{tab:computational_times0}
\end{table}

Table \ref{tab:computational_times0} compares the computational costs of CFD numerical simulations, 
FlowBERT training, and POD-Transformer training. Our model achieves comparable accuracy while 
requiring only 1/9 of the CFD simulation time and half the training time of the POD-Transformer approach.

\subsection{Transfer under Low-Speed Airfoil Conditions}
\vspace{1em}

\begin{figure}[htbp]
	\centering
	\includegraphics[width=\linewidth]{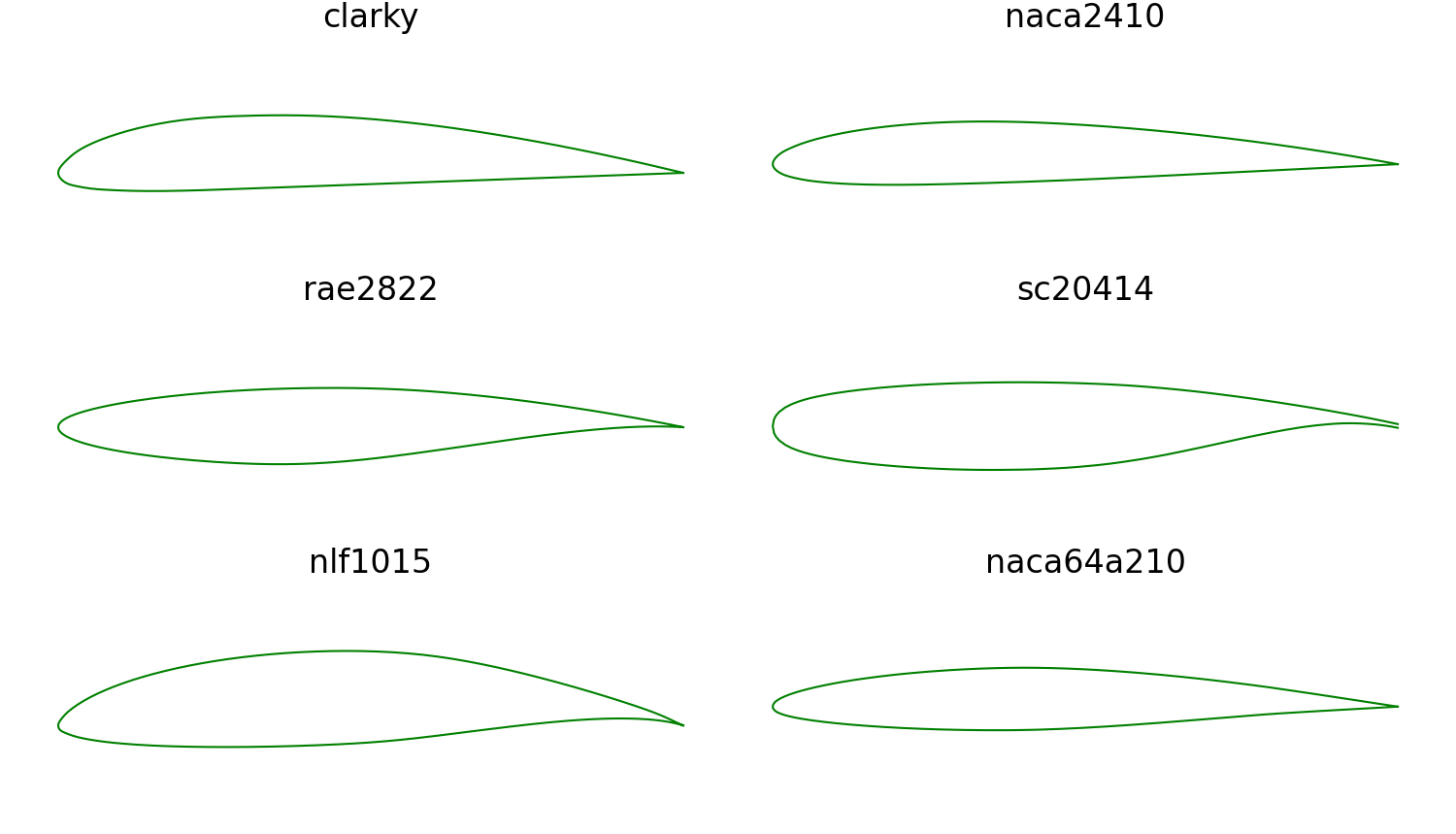}
	\caption{Airfoil Comparison Plot}
	\label{FIG.10}
\end{figure}

\begin{table*}[htbp]
  \centering
  \renewcommand{\arraystretch}{1.5}
  \caption{Airfoil dataset}
  \label{airfoil dataset}
  \begin{tabular*}{\textwidth}{l@{\extracolsep{\fill}} p{3cm} p{3cm} c c}
      \hline
      & \textbf{Airfoil Name} & \textbf{Angle of Attack} & \textbf{Mach Number} & \textbf{Reynolds Number} \\
      \hline
      \multirow{2}{*}{Low speed airfoils} & \multirow{2}{*}{Clarks / NACA2410} & \multirow{2}{*}{0, 4, 8} & 0.2 & $5 \times 10^6$ \\
      \cline{4-5}
      & & & 0.6 & $6.5 \times 10^6$ \\ 
      \hline
      \multirow{2}{*}{Supercritical airfoils} & \multirow{2}{*}{RAE2822 / SC(2)-0414} & \multirow{2}{*}{0, 4, 8} & 0.2 & $5 \times 10^6$ \\
      \cline{4-5}
      & & & 0.6 & $6.5 \times 10^6$ \\ 
      \hline
      \multirow{2}{*}{Laminar airfoils} & \multirow{2}{*}{NLF1015 / NACA64A210} & \multirow{2}{*}{0, 4, 8} & 0.2 & $5 \times 10^6$ \\
      \cline{4-5}
      & & & 0.6 & $6.5 \times 10^6$ \\ 
      \hline
  \end{tabular*}
\end{table*}

In this subsection, we select two low-speed airfoils—the NACA2410 and Clarky airfoils—for 
experimental investigation. These two datasets are used for training and transfer extrapolation 
in our model. As shown in table \ref{airfoil dataset}, our dataset consists of six flow field cases obtained through 
numerical simulation, covering both airfoils at Mach numbers of 0.2 and 0.6 with angles of attack 
of 0°, 4°, and 8°. The airfoil comparison is shown in Fig. {\ref{FIG.10}}.
To minimize the time cost of data acquisition, we collected only 110 flow field 
snapshots for each case, which capture the complete evolution from initial flow development to convergence.

During the numerical simulations using Fluent, we divided the computational domain into 19,274 nodes 
and applied grid refinement around the airfoils. All data for the airfoils were obtained through numerical 
simulations under the same conditions. In this experiment, we plan to use the six flow field datasets 
from the NACA2410 airfoil to train the model, followed by extrapolation tests of its predictive 
capabilities on the Clark-Y airfoil dataset.
Specifically, we used the first 80\% of each flow field dataset (time steps from 1 to 88) as the 
training set for the model training process. The remaining 20\% (time steps from 88 to 110) was 
designated as the validation set to assess the model's predictive performance.

In this experiment, we constructed ROMs for fluid density and pressure. After training the model 
with the NACA2410 airfoil data, we will validate the model using the Clark-Y airfoil data. 
We take the first 10\% of the flow field data from the Clark-Y airfoil and perform autoregressive 
extrapolation to 90 time steps on the trained model (where the simulation data typically reaches a 
convergent state). The time window is set to 10.

Before training the model, it is essential to select an appropriate number of basis functions. 
If the number of basis functions is too low, it may reduce the accuracy of the model's predictions, 
while increasing the number of basis functions can capture richer information, thus reducing the 
reconstruction error of the flow field. However, as the number of basis functions increases, the 
rate of error reduction gradually slows down, 
indicating that simply adding more basis functions may not significantly enhance model accuracy 
and could lead to increased computational costs. Given that the airfoil dataset has 19,274 grid 
points, all experiments in this section and the next will use 18 basis functions as the benchmark.

\begin{figure}
	\centering
	\includegraphics[width=\columnwidth]{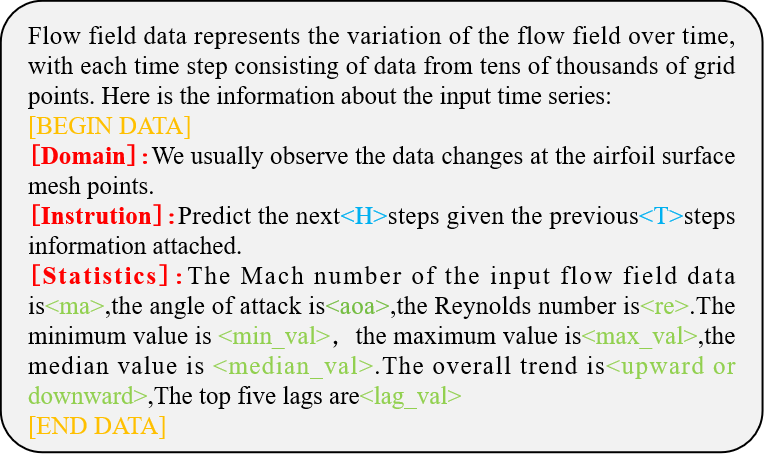}
	\caption{Airfoil text prompt template.}
	\label{FIG.11}
\end{figure}

Before our flow field data enters the model for prediction, we first generate a text prompt 
template for the airfoil flow field data sequence, as shown in Fig. {\ref{FIG.11}}. For each 
time series (ten timesteps), the model generates a text prompt template to guide its predictions. 
Since the flow dynamics of the airfoil vary under different Mach numbers and angles of attack, 
the text prompt template provides a mechanism for the model to better identify and predict the 
flow changes under these varying conditions. By embedding parameter information in the template, 
the model can more effectively learn the temporal evolution patterns of the flow field, thereby 
enhancing prediction accuracy and reliability.

\begin{figure*}
	\centering
	\includegraphics[width=.9\textwidth]{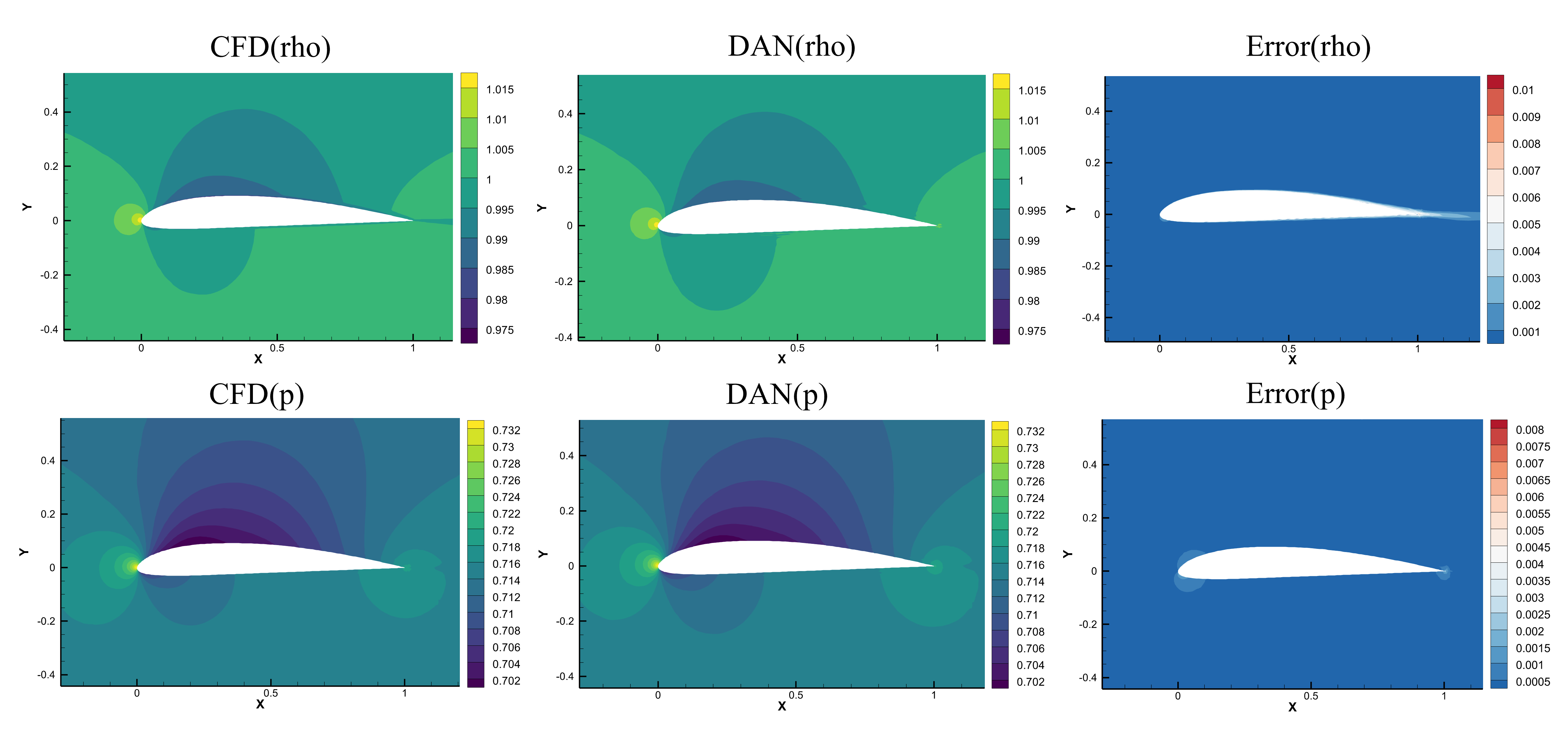}
	\caption{Extrapolated density field (rho) and pressure field (p) for the Clark-Y airfoil at Ma=0.2, AOA=0°, and Re=5e6 reaching convergence.}
	\label{FIG.12}
\end{figure*}

\begin{figure*}
  \centering
  \begin{subfigure}{0.45\textwidth}
      \includegraphics[width=\linewidth]{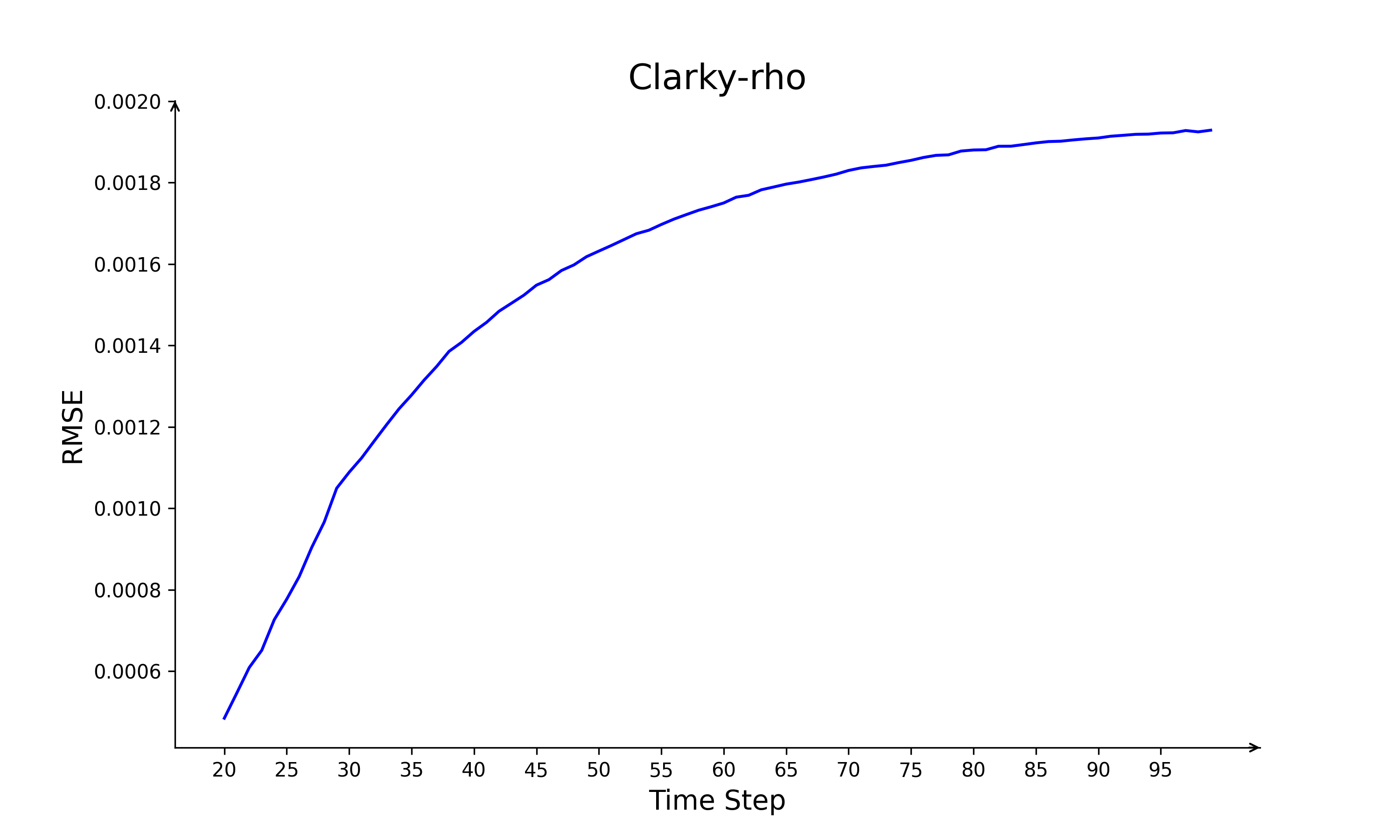}
      \label{FIG.13a}
  \end{subfigure}
  \hspace{0.01\textwidth}
  \begin{subfigure}{0.45\textwidth}
      \includegraphics[width=\linewidth]{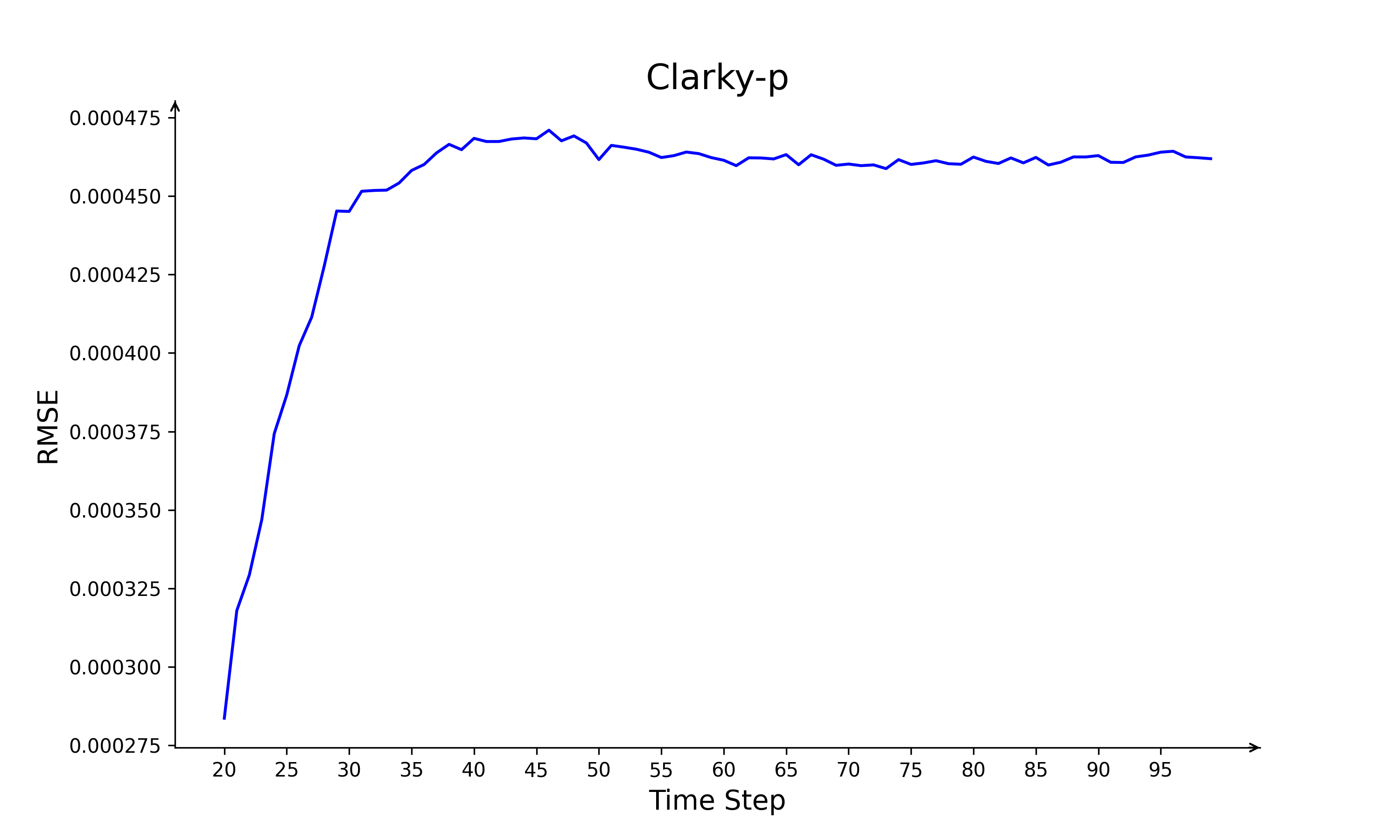}
      \label{FIG.13b}
  \end{subfigure}
  \caption{Extrapolation error rmse for the Clark-Y airfoil at Ma=0.2, AOA=0°, and Re=5e6.}
  \label{FIG.13}
\end{figure*}

\begin{figure*}
	\centering
	\includegraphics[width=.9\textwidth]{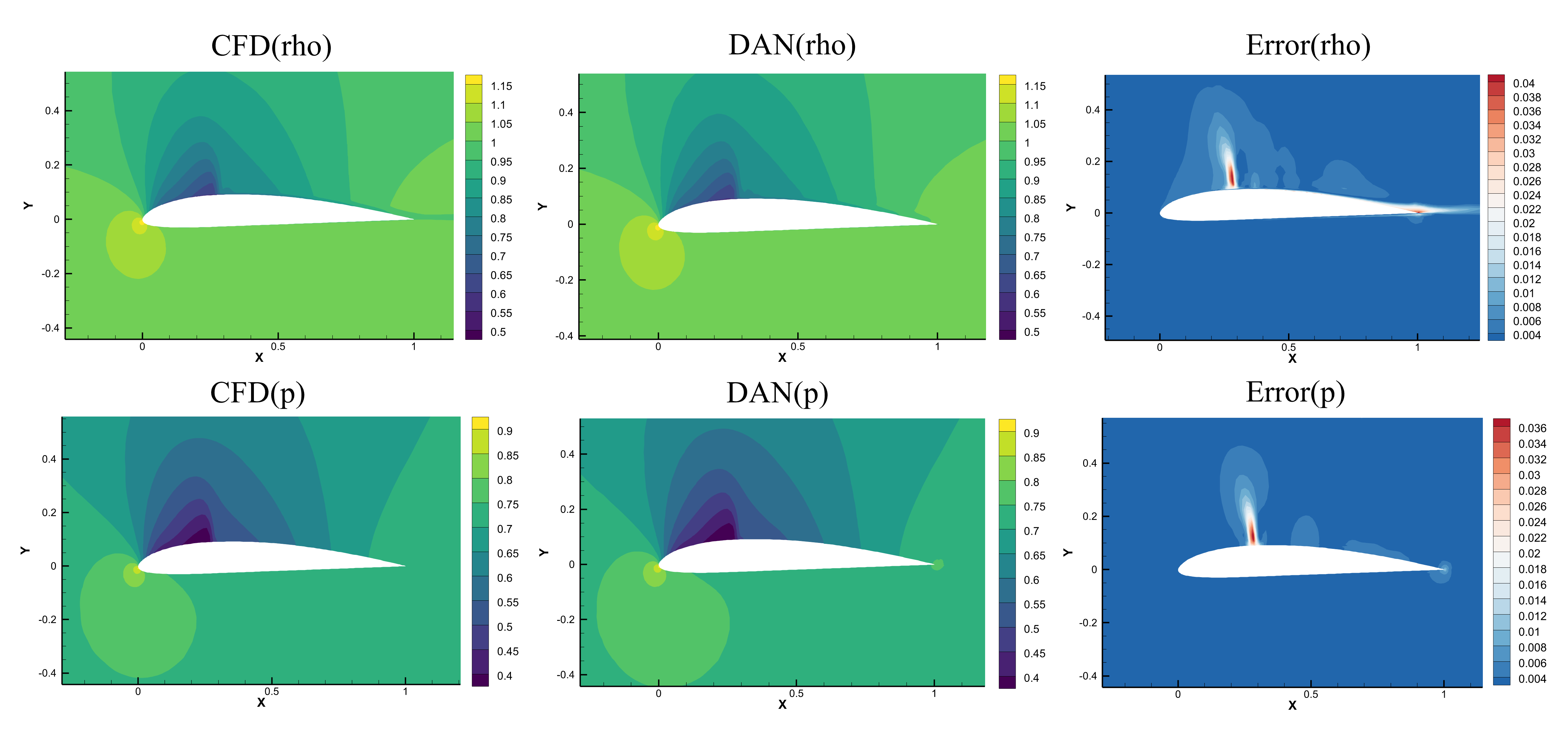}
	\caption{Extrapolated density field (rho) and pressure field (p) for the Clark-Y airfoil at Ma=0.6, AOA=4°, and Re=6.5e6 reaching convergence.}
	\label{FIG.14}
\end{figure*}

\begin{figure*}
  \centering
  \begin{subfigure}{0.45\textwidth}
      \includegraphics[width=\linewidth]{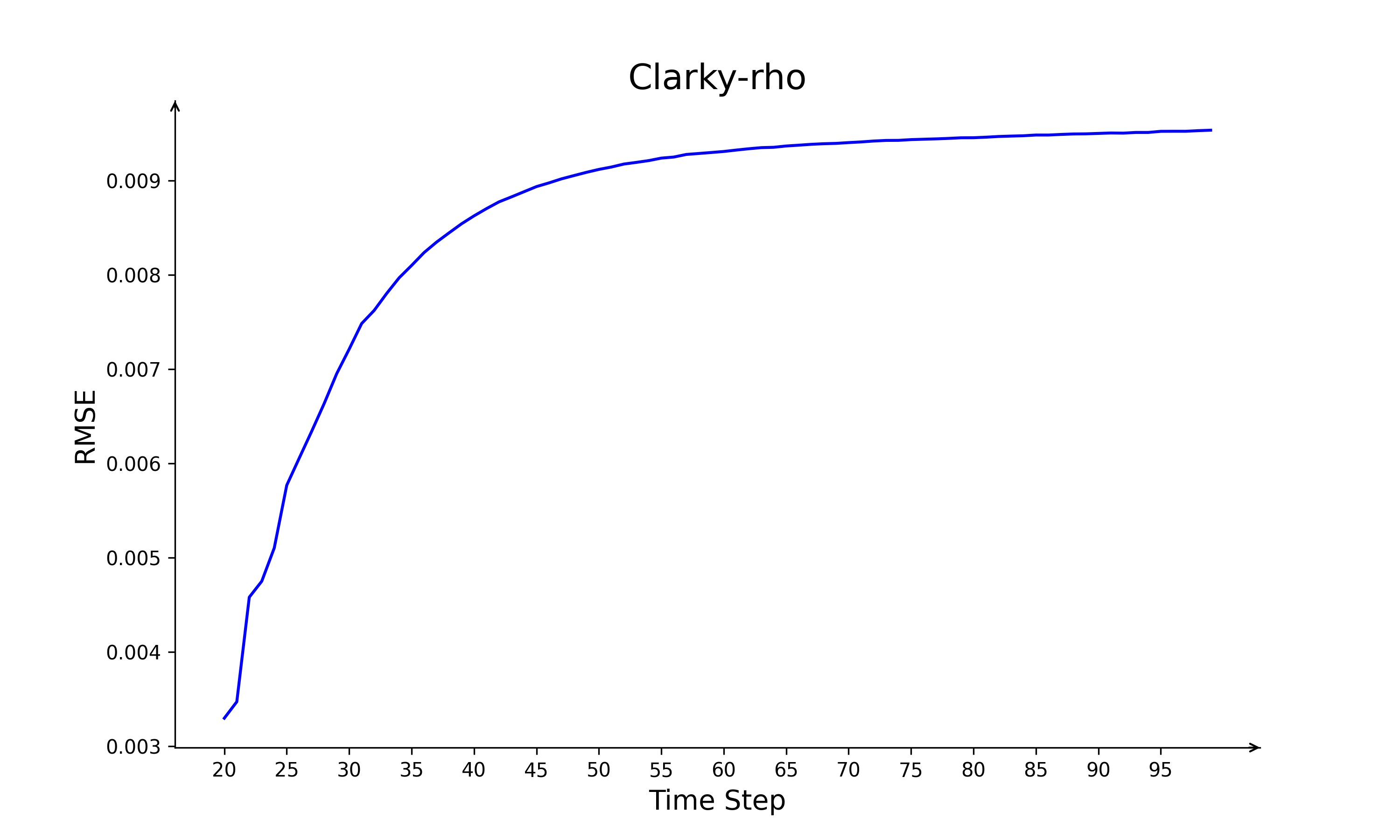}
      \label{FIG.15a}
  \end{subfigure}
  \hspace{0.01\textwidth}
  \begin{subfigure}{0.45\textwidth}
      \includegraphics[width=\linewidth]{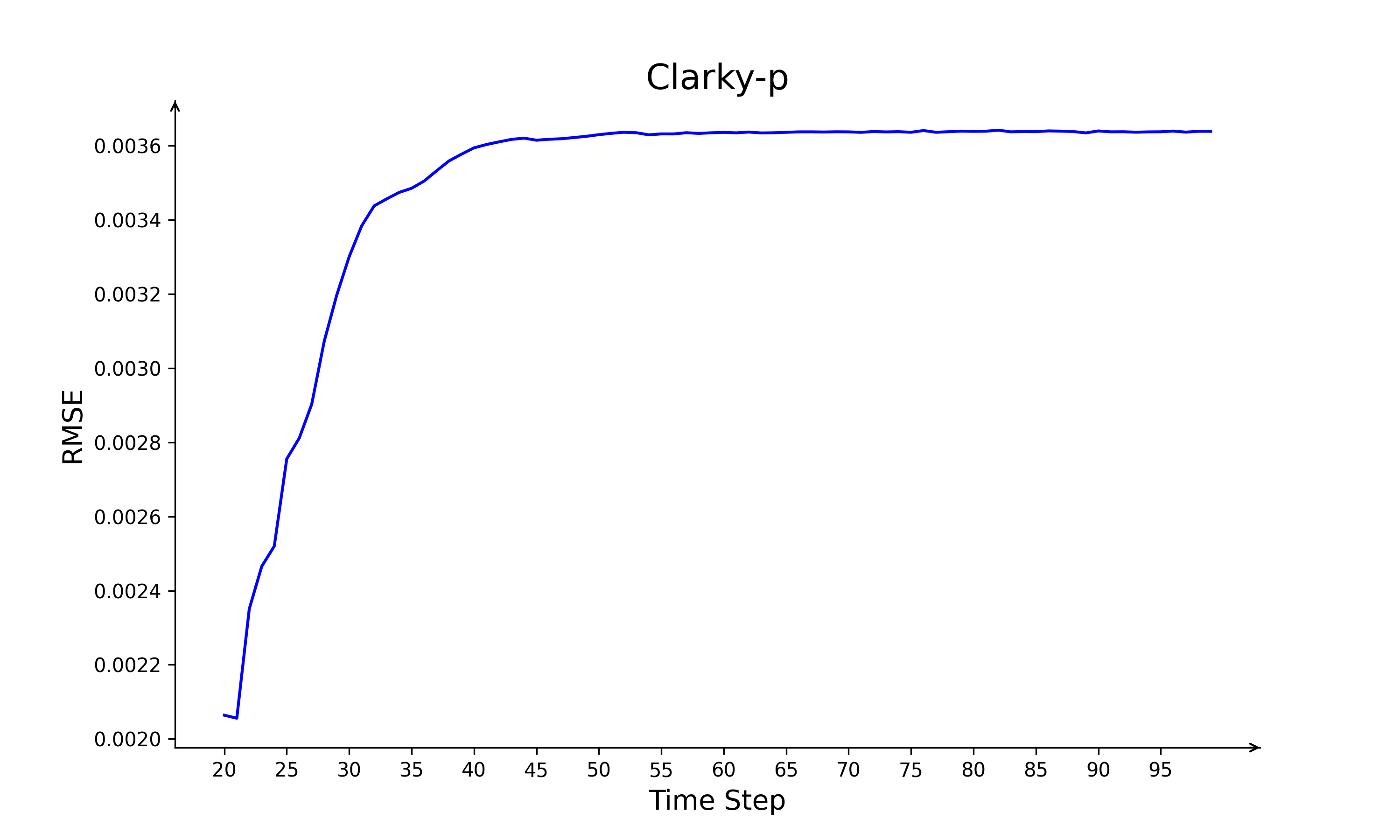}
      \label{FIG.15b}
  \end{subfigure}
  \caption{Extrapolation error rmse for the Clark-Y airfoil at Ma=0.6, AOA=4°, and Re=6.5e6.}
  \label{FIG.15}
\end{figure*}

During the model validation phase, we selected flow field data from the Clark-Y airfoil at 
Ma = 0.2, AOA = 0°, and Ma = 0.6, AOA = 4° for testing; these datasets were 
not used in model training. 
Fig. {\ref{FIG.12}} and Fig. {\ref{FIG.14}} illustrate the comparison of CFD results with 
model predictions and error analysis for these conditions. The results show that, under low 
Mach numbers and angles of attack, the model's predicted density field matches well with CFD 
results, with a maximum absolute error of 0.0075, 
primarily occurring near the wall and in areas of rapid change at the leading edge. The 
pressure field predictions near the airfoil boundaries also maintained high accuracy, with a 
maximum error of 0.0045. Overall, the prediction results exhibit high accuracy. When the Mach number and angle of attack 
increase to 0.6 and 4, respectively, the model still performs well in predicting both the density 
and pressure fields, consistent with CFD results. However, the error maps show that the maximum 
errors for the density and pressure fields rise to 0.042 and 0.036, concentrated at the leading edge. 
This increase in error is expected, as higher Mach numbers and angles of attack accelerate flow 
convergence and change trends more rapidly, reducing the effective data available for training 
from our 110 time steps and consequently decreasing the amount of information the model can learn, 
leading to larger extrapolation errors. In summary, our model continues to demonstrate outstanding 
performance in transfer learning.

We also conducted training and extrapolation experiments on the Transformer-based ROM. The experimental 
results reveal that this model can only achieve accurate predictions under a single flow condition. When 
the flow conditions change and require transfer extrapolation, the model fails to maintain stability, 
exhibiting significant prediction errors that extend to even fundamental flow characteristics.

In addition to analyzing the convergent state flow field maps and absolute error maps obtained 
from the model extrapolation, we also plotted the rmse curves throughout the extrapolation process. 
Fig. {\ref{FIG.13}} and Fig. {\ref{FIG.15}} illustrate the rmse curves for the model extrapolation 
under conditions of Ma =  0.2, AOA = 0 and Ma = 0.6, AOA = 4, respectively. 
From the curves, it is evident that errors gradually accumulate during the extrapolation process, 
but they tend to stabilize and no longer significantly increase as the convergence state approaches. 
This indicates that, regardless of the conditions, our model can effectively capture the flow field 
trends of the airfoil and identify the convergent state. These results align with our expectations 
for the model and open up possibilities for exploring a wider range of different flow field 
conditions in the future.

The research results demonstrate that the trained flow field prediction model exhibits exceptional 
extrapolation capabilities, validating the effectiveness of our proposed text prompt templates and 
temporal text embedding strategies. By inputting the parameter conditions of the airfoil flow field 
into the text prompt templates and converting the sequential flow field data into text format, we 
utilized a pre-trained large model for temporal prediction, achieving precise flow field predictions.

Table \ref{tab:computational_times1} compares the computational time required for network training 
and prediction with that of CFD simulations. Our method significantly enhances speed compared to 
traditional CFD approaches, where the CFD computation for a single airfoil under six conditions 
takes 8,280 seconds. In contrast, the training phase requires only 900 seconds, and once training 
is complete, the autoregressive extrapolation process takes just 10 seconds to reach a convergent 
state for a single airfoil flow field. In most cases, fine-tuning the airfoil typically necessitates 
recalculating the entire flow field, whereas our model can directly extrapolate to the convergent 
state by computing only the first 10\% of the flow field data, greatly improving efficiency.

\begin{table}[ht]
  \centering
  \caption{Computational times for numerical simulations and ROM training and extrapolation}
  \begin{tabular}{@{}p{0.35\textwidth} p{0.1\textwidth}@{}}
      \toprule
      Method & Time \\ \midrule
      Numerical Simulation & 8280 s \\ 
      ROM Training Time & 900 s \\ 
      ROM Extrapolation Time & 10 s \\ 
      \bottomrule
  \end{tabular}
  \label{tab:computational_times1}
\end{table}

\subsection{Transfer under multiple airfoil conditions}
\vspace{1em}

In this part of the study, we will investigate flow field data from different types of airfoils 
under various conditions. As shown in Table \ref{airfoil dataset}, we categorize the airfoils into three main types: 
low-speed airfoils, supercritical airfoils, and laminar airfoils, with each category containing 
two different designs. The flow field data for each airfoil is divided into six subsets based on 
different Mach numbers and angles of attack.

We plan to use the flow field data from the NACA2410, RAE2822, and NLF(1)-015 airfoils for model 
training, while the data from the Clark-Y, NACA64A210, and SC(2)-0414 airfoils will be used for 
model validation. The validation process will follow the same method described in the previous section, 
where we perform autoregressive predictions using the first 10\% of the flow field data and 
extrapolate to the convergent state at the 90th time step. We will evaluate the model's performance 
by comparing the predicted convergent state with the actual results through visual representations 
and analyzing the absolute error maps between the two.

In this experiment, we also constructed ROM for fluid density and pressure, using 18 basis 
functions as the benchmark. In the previous section, we validated that our ROM demonstrates good 
transferability even when the airfoil shape changes within the same category. 
In this section, we will further verify through experiments that our reduced-order model can achieve 
the expected performance and accuracy even when dealing with different categories of airfoils and 
varying shapes.

Following the same approach as in the previous section, for each set of flow field data fed into 
the model, we will create a corresponding text description template, as shown in Fig. {\ref{FIG.11}}. 
For each time series data, we will design a template that 
includes parameter information specific to the airfoil flow field and data features derived from 
analytical computations. After processing by a pre-trained language model, these templates will be 
used to guide the model's prediction process for the flow field data.

\begin{figure*}
	\centering
	\includegraphics[width=.9\textwidth]{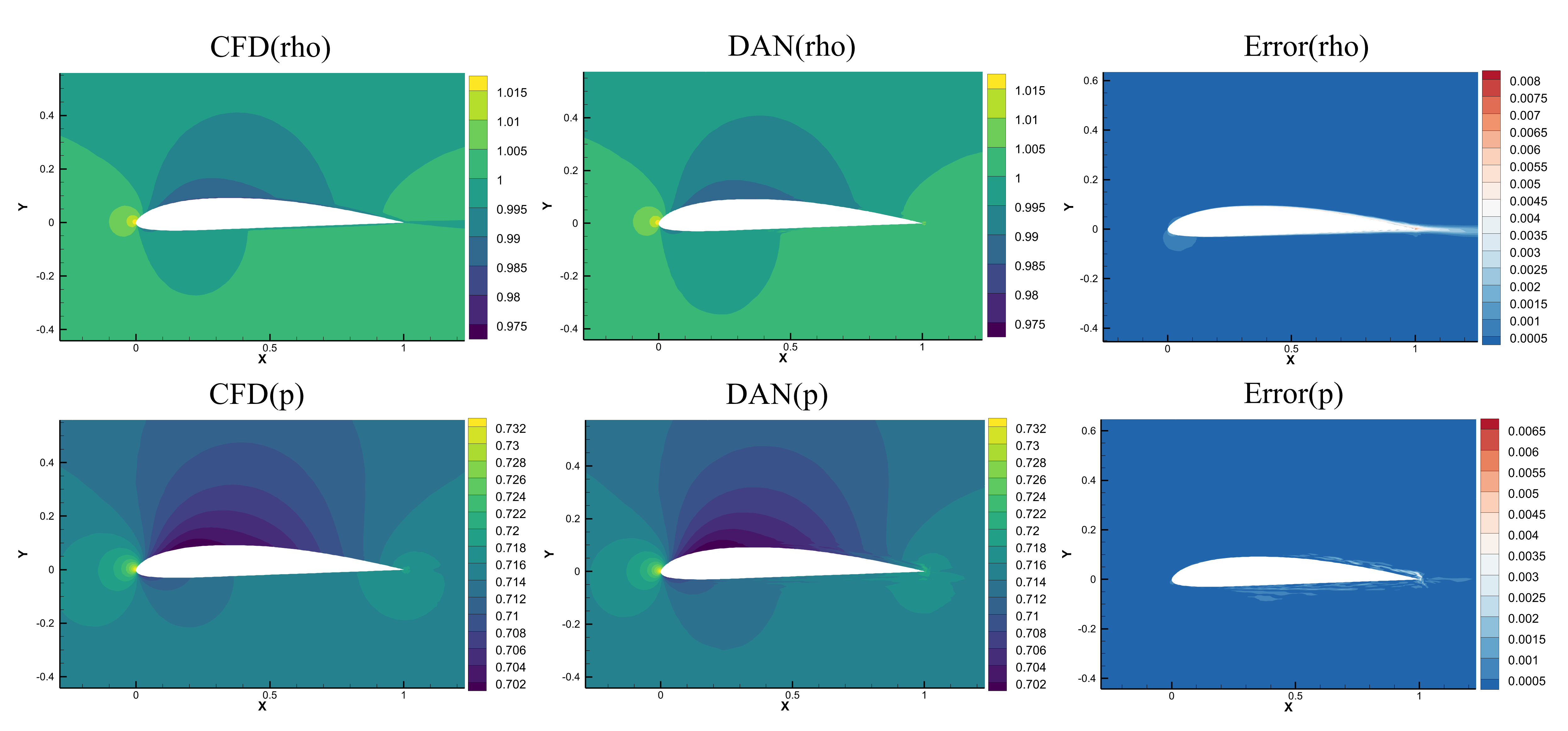}
	\caption{Extrapolated density field (rho) and pressure field (p) for the Clark-Y airfoil at Ma=0.2, AOA=0°, and Re=5e6 reaching convergence.}
	\label{FIG.16}
\end{figure*}

During the model validation phase, we first filtered the parameters of the airfoil flow field data 
used for validation. For the low-speed airfoil Clark-Y, we selected the flow field data at Ma = 0.2 
and AOA = 0° for extrapolation validation. 
Fig. {\ref{FIG.16}} shows the comparison of the model's extrapolated density field and pressure field 
with the CFD results under these conditions, along with the absolute error map between the two in the 
convergent state.
Comparative analysis reveals that both the density and pressure field predictions closely match the 
CFD results, with no significant errors observed. There are some minor discrepancies present on the 
upper and lower surfaces of the airfoil and in the trailing edge region.

\begin{figure*}
  \centering
  \begin{subfigure}{0.45\textwidth}
      \includegraphics[width=\linewidth]{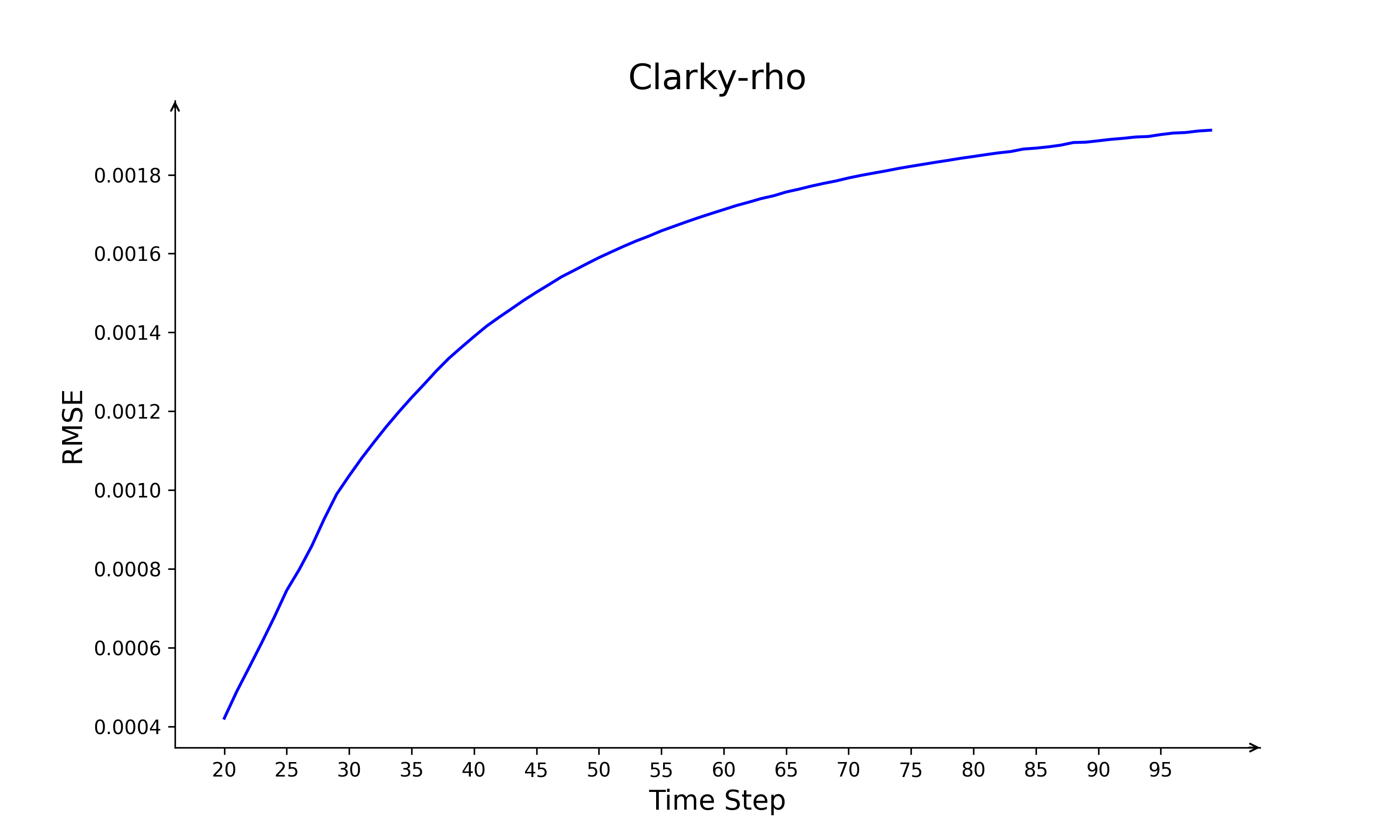}
      \label{FIG.17a}
  \end{subfigure}
  \hspace{0.01\textwidth}
  \begin{subfigure}{0.45\textwidth}
      \includegraphics[width=\linewidth]{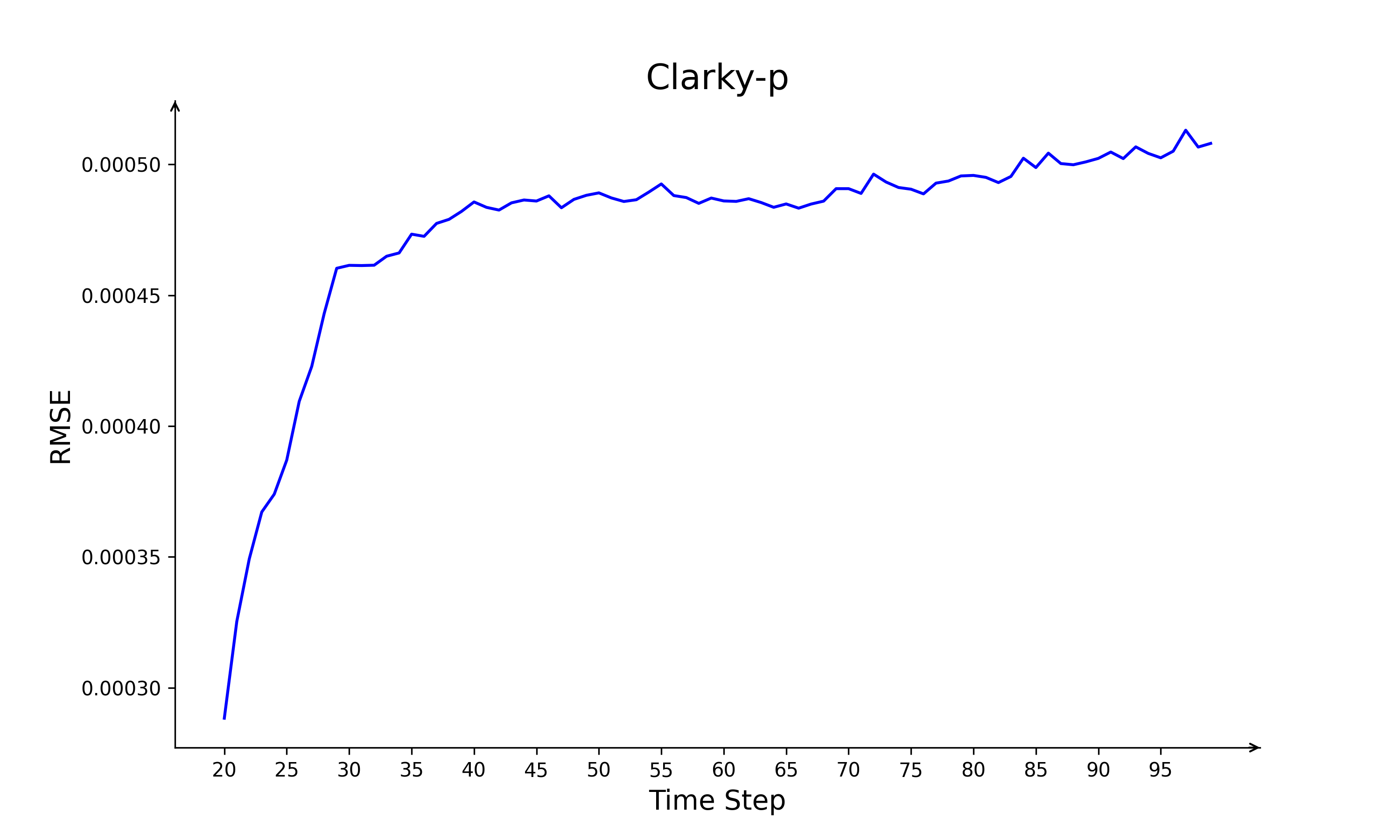}
      \label{FIG.17b}
  \end{subfigure}
  \caption{Extrapolation error rmse for the Clark-Y airfoil at Ma=0.2, AOA=0°, and Re=5e6.}
  \label{FIG.17}
\end{figure*}

\begin{figure*}
	\centering
	\includegraphics[width=.9\textwidth]{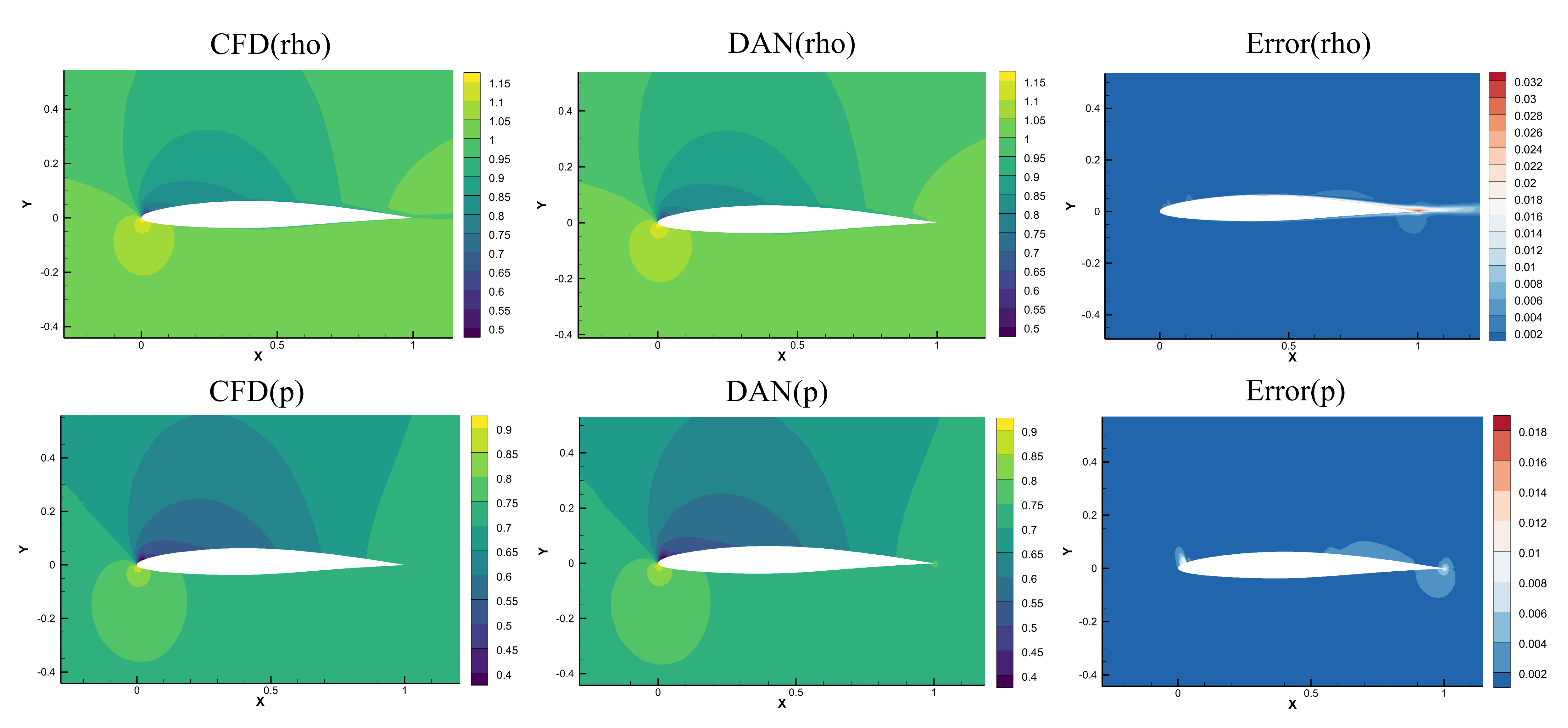}
	\caption{Extrapolated density field (rho) and pressure field (p) for the NACA64A210 airfoil at Ma=0.6, AOA=4°, and Re=6.5e6 reaching convergence.}
	\label{FIG.18}
\end{figure*}

\begin{figure*}
  \centering
  \begin{subfigure}{0.45\textwidth}
      \includegraphics[width=\linewidth]{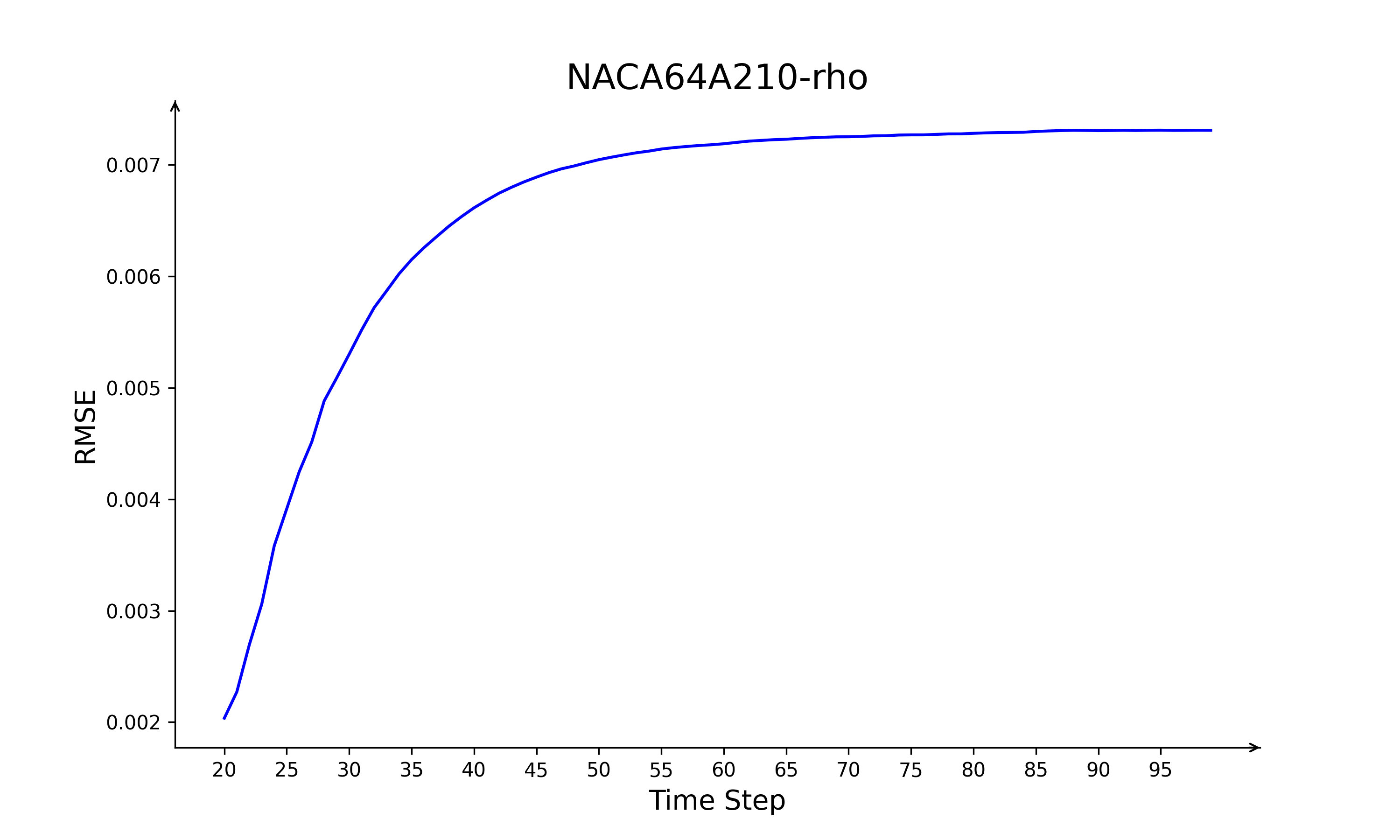}
      \label{FIG.19a}
  \end{subfigure}
  \hspace{0.01\textwidth}
  \begin{subfigure}{0.45\textwidth}
      \includegraphics[width=\linewidth]{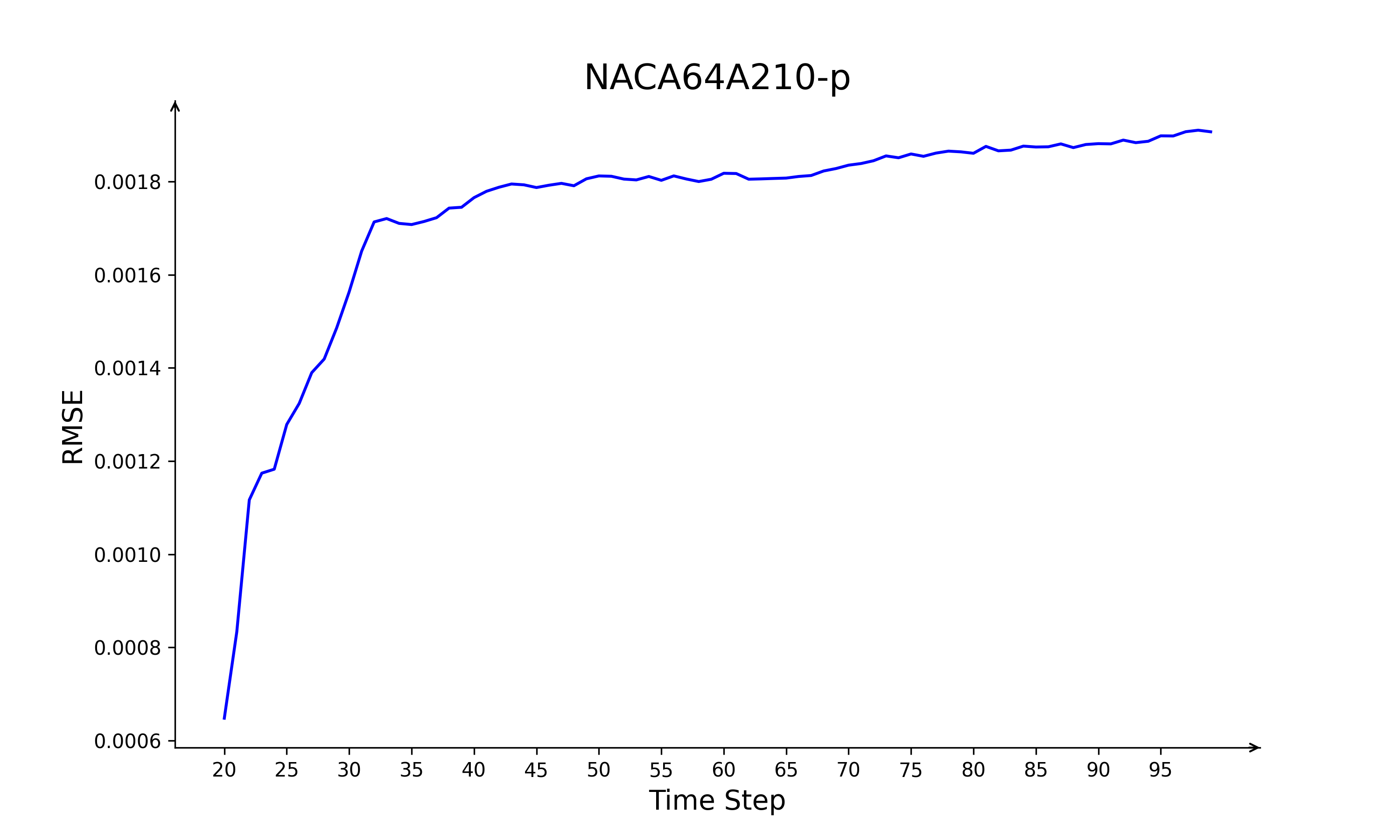}
      \label{FIG.19b}
  \end{subfigure}
  \caption{Extrapolation error rmes for the NACA64A210 airfoil at Ma=0.6, AOA=4°, and Re=6.5e6.}
  \label{FIG.19}
\end{figure*}

\begin{figure*}
	\centering
	\includegraphics[width=.9\textwidth]{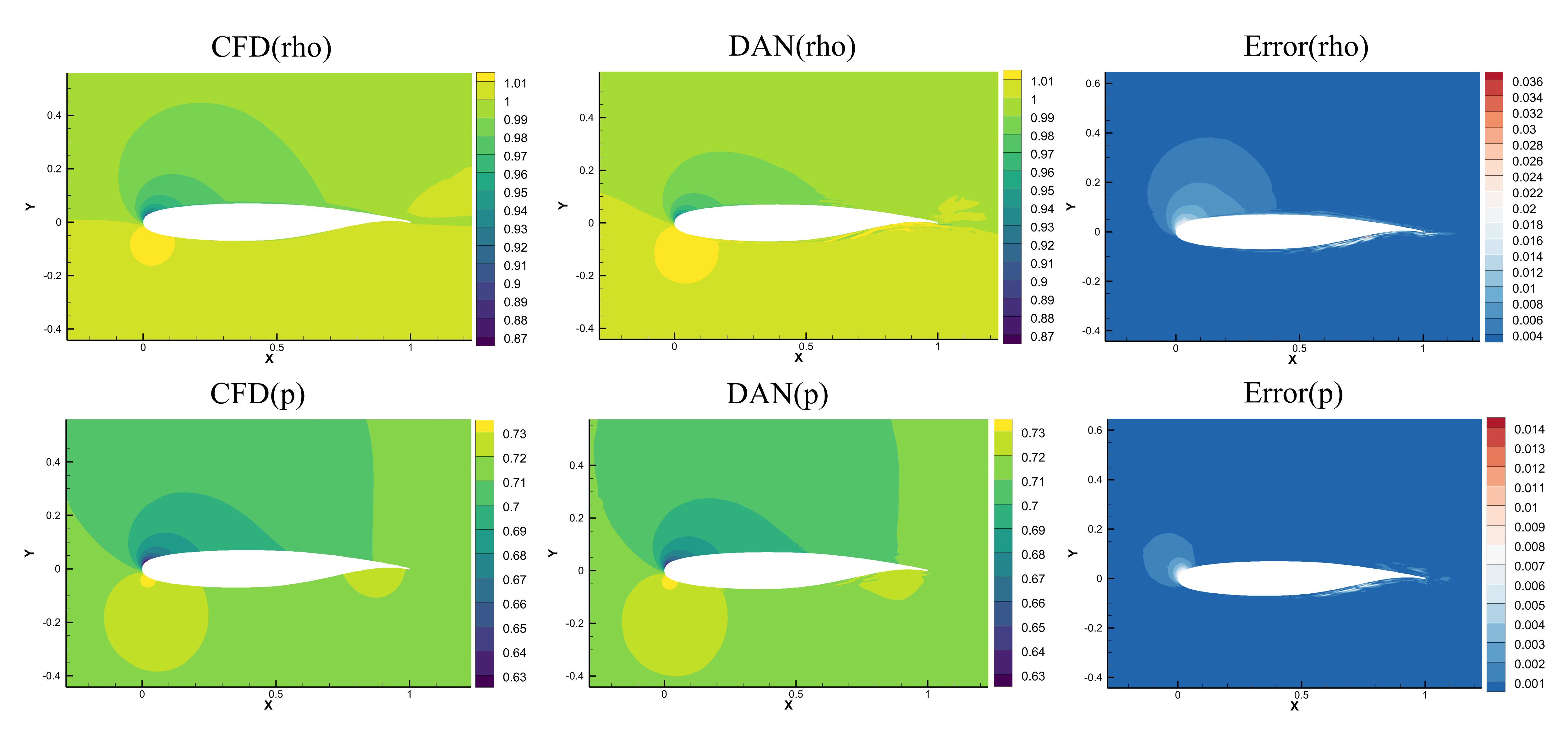}
	\caption{Extrapolated density field (rho) and pressure field (p) for the SC(2)-0414 airfoil at Ma=0.2, AOA=8°, and Re=5e6 reaching convergence.}
	\label{FIG.20}
\end{figure*}

\begin{figure*}
  \centering
  \begin{subfigure}{0.45\textwidth}
      \includegraphics[width=\linewidth]{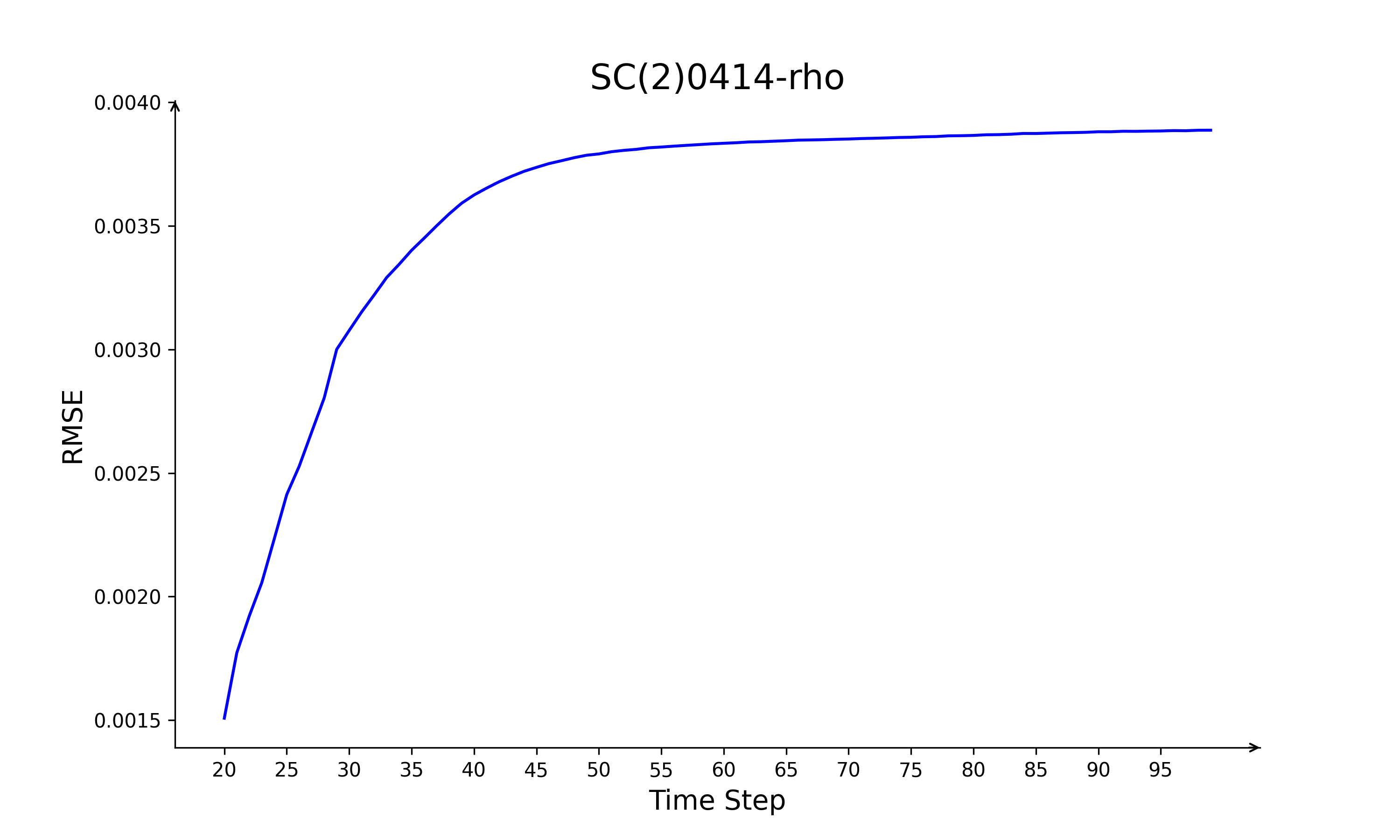}
      \label{FIG.21a}
  \end{subfigure}
  \hspace{0.01\textwidth}
  \begin{subfigure}{0.45\textwidth}
      \includegraphics[width=\linewidth]{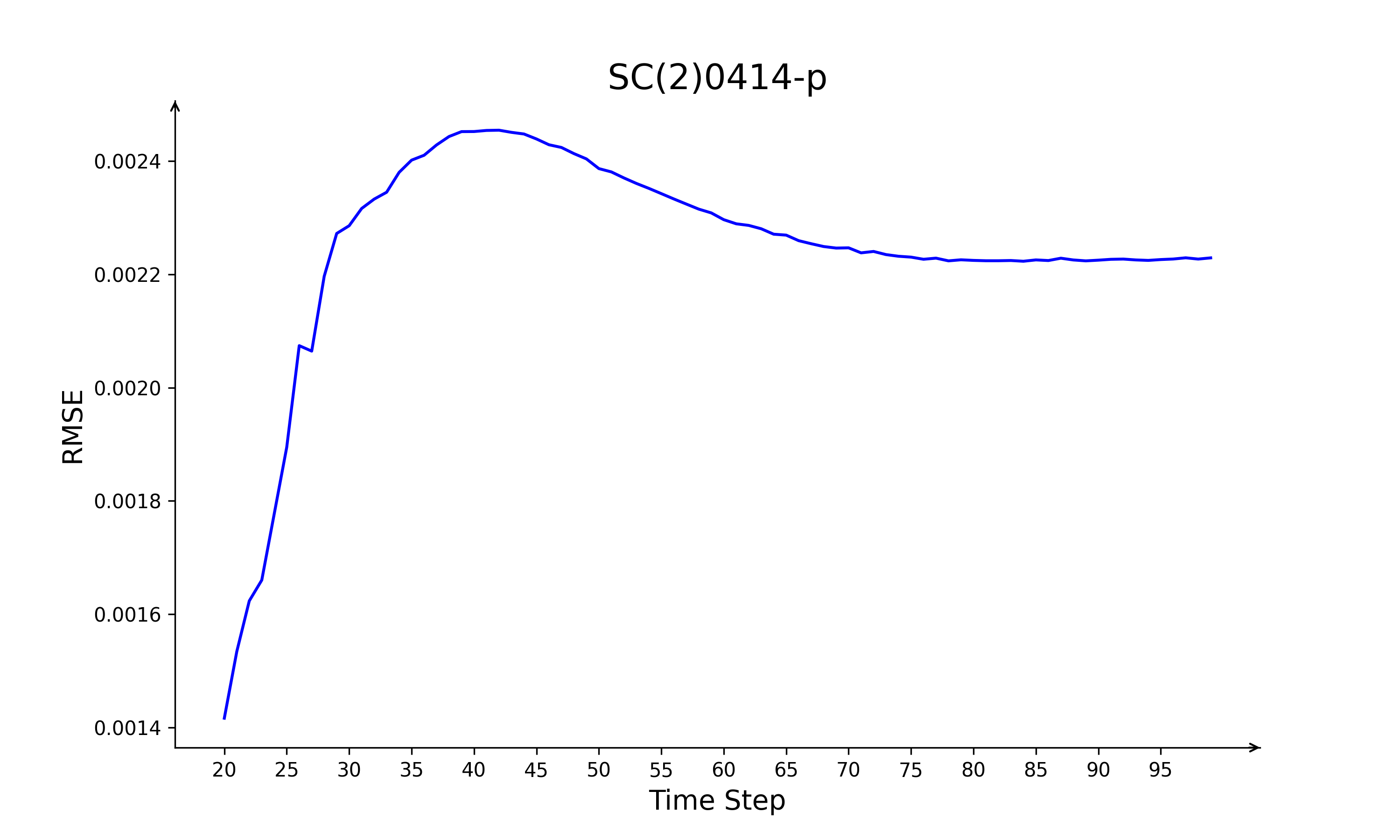}
      \label{FIG.21b}
  \end{subfigure}
  \caption{Extrapolation error rmse for the SC(2)-0414 airfoil at Ma=0.2, AOA=8°, and Re=5e6.}
  \label{FIG.21}
\end{figure*}

From the absolute error map, it is evident that the error distribution aligns largely with our previous 
expectations. Specifically, the maximum error for the density field is 0.008, while the maximum error 
for the pressure field is 0.0065. 
Compared to the results from the previous section, the errors have increased. This change is attributed 
to the doubling of the data volume used for model training in this section, which heightened the model's 
difficulty in capturing specific flow field variations of the airfoil, 
particularly in regions of rapid change such as the upper and lower surfaces and the trailing edge. 
Nonetheless, the errors remain at a low level. Therefore, we can conclude that the model's performance 
in transfer learning for low-speed airfoils is quite satisfactory.

For the laminar airfoil NACA64A210, we selected flow field data at Ma = 0.6 and AOA = 4° for 
extrapolation validation. Fig. {\ref{FIG.18}} shows the comparison between 
the model predictions for the density field and pressure field with the CFD results, along 
with the corresponding error distribution. 
The results indicate that the model predictions are generally consistent with the CFD results, 
with only slight errors observed at the trailing edge.The absolute error map reveals that the 
density field errors are primarily concentrated at the trailing edge, 
while the pressure field errors are distributed in both the trailing edge and near-wall regions, 
with maximum errors of 0.032 and 0.013, respectively. As the Mach number and angle of attack 
increase, the model prediction errors also rise, attributed to the accelerated changes in the 
flow field, which reduce the amount of time step information available for convergence, thereby 
affecting prediction accuracy. Nevertheless, the model's extrapolation performance under these 
conditions remains good.

For the supercritical airfoil SC(2)-0414, we conducted extrapolation validation of the flow field 
data at Ma = 0.2 and AOA = 8°. Fig. {\ref{FIG.20}} shows that the model predictions 
for the density field and pressure field are in good agreement with the CFD results, with no 
significant errors observed.
The absolute error map indicates that, with the increase in angle of attack, the prediction errors 
at the leading edge of the airfoil are more pronounced, and there are also some discrepancies on 
the upper and lower surfaces. The maximum errors for the density field and pressure field are 
0.034 and 0.014, respectively. 
Although these values are higher than those under previous conditions, the overall performance 
remains satisfactory.

Fig. {\ref{FIG.17}}, Fig. {\ref{FIG.19}} and Fig. {\ref{FIG.21}}  display the RMSE curves for 
extrapolation under three different airfoil conditions. Observing these curves reveals that, 
regardless of whether at low or high Mach numbers, the RMSE eventually stabilizes, indicating 
that the model can effectively predict the convergent state. 
The average RMSEs under these three conditions are as follows: 0.004 for Clarky, 0.006 for 
NACA64A210, and 0.01 for SC(2)-0414. These error values fall within the expected range, 
demonstrating the model's excellent performance in transfer learning and extrapolation, 
as well as its strong ability to capture flow field variations.
Table \ref{tab:computational_times2} lists the time required to train the models for the 
three airfoils, the time to extrapolate a flow field for one airfoil, and the time required 
for numerical simulations. The data indicate that CFD calculations for 18 flow field datasets 
take 24,840 seconds, while training for these datasets only requires 2,700 seconds, 
and the model can extrapolate a flow field in just 10 seconds. This highlights the 
significant time efficiency of our model. Overall, the proposed ROM achieves good transferability 
across various airfoil conditions, effectively addressing the issue of needing to retrain 
the model due to adjustments or changes in airfoil design.

\begin{table}[ht]
  \centering
  \caption{Computational times for numerical simulations and ROM training and extrapolation.}
  \begin{tabular}{@{}p{0.35\textwidth} p{0.1\textwidth}@{}}
      \toprule
      Method & Time \\ \midrule
      Numerical Simulation & 24840 s \\ 
      ROM Training Time & 2700 s \\ 
      ROM Extrapolation Time & 10 s \\ 
      \bottomrule
  \end{tabular}
  \label{tab:computational_times2}
\end{table}

\section{Conclusion}
\vspace{1em}

This paper proposes a reduced-order model  that integrates Proper Orthogonal Decomposition 
with pretrained large language models. The model employs POD to extract critical low-dimensional 
features from high-fidelity numerical simulations, followed by leveraging pretrained LLM to uncover 
temporal dynamics within these low-dimensional features. Prior to utilizing the pretrained LLM for 
prediction, we developed a text template specialized for flow field time-series data and an embedding 
technique to convert low-dimensional temporal flow data into text vectors. The compiled text template 
guides the LLM in temporal prediction, while the embedding technique ensures the low-dimensional flow 
data is formatted in a manner interpretable to the pretrained LLM. This approach significantly enhances 
the performance of pretrained LLM in flow field time-series prediction and enables potential transfer 
learning across diverse scenarios. Although the pretrained LLM applied here is relatively small-scale, 
the recent proliferation of large-scale pretrained models suggests that integrating them into our 
architecture could yield superior performance, while our framework still achieves desired results with 
minimal training epochs.


\section*{CRediT authorship contribution statement}
\vspace{1em}
Weihao Zou: Conceptualization, Methodology, Software, Validation, Writing. 
Weibing Feng: Conceptualization, Methodology, Software, Validation, Writing. 
Pin Wu: Conceptualization, Methodology, Software, Writing.
Jiangnan Wu: Methodology, Software, Validation.
Yiguo Yang: Methodology, Software, Validation.

\section*{Declaration of Competing Interest}
\vspace{1em}
The authors declare that they have no known competing financial 
interests or personal relationships that could have appeared to influence 
the work reported in this paper.

\section*{Data available}
\vspace{1em}
The data that support the findings of this study are available
from the corresponding author upon reasonable request.


\printcredits

\bibliographystyle{unsrtnat}

\bibliography{cas-refs}





\end{document}